\documentclass{article}

\usepackage{PRIMEarxiv}

\usepackage[utf8]{inputenc} 
\usepackage[T1]{fontenc}    
\usepackage{hyperref}       
\usepackage{url}            
\usepackage{booktabs}       
\usepackage{amsfonts}       
\usepackage{nicefrac}       
\usepackage{microtype}      
\usepackage{lipsum}
\usepackage{graphicx}
\graphicspath{{media/}}     

\usepackage[utf8]{inputenc} 
\usepackage[T1]{fontenc}    
\usepackage{hyperref}       
\usepackage{url}            
\usepackage{booktabs}       
\usepackage{amsfonts}       
\usepackage{nicefrac}       
\usepackage{microtype}      
\usepackage{graphicx}       
\usepackage{bbm}
\usepackage{amsmath}
\usepackage{subcaption}
\usepackage[table]{xcolor}
\usepackage{algorithm}
\usepackage{algpseudocode}
\usepackage{multirow}

\usepackage{amssymb}
\usepackage{pifont}
\newcommand{\cmark}{\ding{51}}%
\newcommand{\xmark}{\ding{55}}%

\usepackage{wrapfig}
\usepackage{colortbl}
\graphicspath{{media/}}     
\newcommand{\wjh}[1]{}
\newcommand{\ml}[1]{}

\usepackage{natbib}

\newcommand{\squishlist}{
\begin{list}{{{\small{$\bullet$}}}}
{\setlength{\itemsep}{3pt}      \setlength{\parsep}{1pt}
\setlength{\topsep}{1pt}       \setlength{\partopsep}{0pt}
\setlength{\leftmargin}{1em} \setlength{\labelwidth}{1em}
\setlength{\labelsep}{0.5em} } }
\newcommand{\squishend}{  \end{list}  }

\definecolor{lightblue}{RGB}{0, 102, 204}
\definecolor{blue}{RGB}{0, 76, 153}
\definecolor{lightgreen}{RGB}{34,139,34}
\definecolor{darkred}{RGB}{139,0,0}

\definecolor{codegreen}{rgb}{0,0.6,0}
\definecolor{codegray}{rgb}{0.5,0.5,0.5}
\definecolor{codepurple}{rgb}{0.58,0,0.82}
\definecolor{backcolour}{rgb}{0.95,0.95,0.92}

\usepackage{listings}
\usepackage{xcolor}

\definecolor{codegreen}{rgb}{0,0.6,0}
\definecolor{codegray}{rgb}{0.5,0.5,0.5}
\definecolor{codepurple}{rgb}{0.58,0,0.82}
\definecolor{backcolour}{rgb}{0.95,0.95,0.92}

\lstdefinestyle{mystyle}{
    backgroundcolor=\color{backcolour},   
    commentstyle=\color{codegreen},
    keywordstyle=\color{magenta},
    numberstyle=\tiny\color{codegray},
    stringstyle=\color{codepurple},
    basicstyle=\ttfamily\footnotesize,
    breakatwhitespace=false,         
    breaklines=true,                 
    captionpos=b,                    
    keepspaces=true,                 
    numbers=left,                    
    numbersep=5pt,                  
    showspaces=false,                
    showstringspaces=false,
    showtabs=false,                  
    tabsize=2
}

\newcommand\YAMLcolonstyle{\color{red}\mdseries}
\newcommand\YAMLkeystyle{\color{black}\bfseries}
\newcommand\YAMLvaluestyle{\color{blue}\mdseries}

\makeatletter

\newcommand\language@yaml{yaml}

\expandafter\expandafter\expandafter\lstdefinelanguage
\expandafter{\language@yaml}
{
  keywords={true,false,null,y,n},
  keywordstyle=\color{darkgray}\bfseries,
  basicstyle=\YAMLkeystyle,                                 
  sensitive=false,
  comment=[l]{\#},
  morecomment=[s]{/*}{*/},
  commentstyle=\color{purple}\ttfamily,
  stringstyle=\YAMLvaluestyle\ttfamily,
  moredelim=[l][\color{orange}]{\&},
  moredelim=[l][\color{magenta}]{*},
  moredelim=**[il][\YAMLcolonstyle{:}\YAMLvaluestyle]{:},   
  morestring=[b]',
  morestring=[b]",
  literate =    {---}{{\ProcessThreeDashes}}3
                {>}{{\textcolor{red}\textgreater}}1     
                {|}{{\textcolor{red}\textbar}}1 
                {\ -\ }{{\mdseries\ -\ }}3,
}

\lst@AddToHook{EveryLine}{\ifx\lst@language\language@yaml\YAMLkeystyle\fi}
\makeatother

\newcommand\ProcessThreeDashes{\llap{\color{cyan}\mdseries-{-}-}}

\title{Automatic Dataset Construction (ADC):\\Sample Collection, Data Curation, and Beyond}

\author{\parbox{\linewidth}{\centering\bfseries
Minghao Liu\textsuperscript{1}\thanks{Correspondence: \texttt{mliu40@ucsc.edu}},
Zonglin Di\textsuperscript{1},
Jiaheng Wei\textsuperscript{1},
Zhongruo Wang\textsuperscript{2},
Hengxiang Zhang\textsuperscript{3},
Ruixuan Xiao\textsuperscript{4} \\
Haoyu Wang\textsuperscript{5},
Jinlong Pang\textsuperscript{1},
Hao Chen\textsuperscript{6},
Ankit Shah\textsuperscript{6},
Hongxin Wei\textsuperscript{3},
Xinlei He\textsuperscript{7} \\
Zhaowei Zhu\textsuperscript{1},
Haobo Wang\textsuperscript{4},
Lei Feng\textsuperscript{8},
Jindong Wang\textsuperscript{9},
James Davis\textsuperscript{1},
Yang Liu\textsuperscript{1} \\[0.4em]
{\mdseries\small
\textsuperscript{1}University of California, Santa Cruz\quad
\textsuperscript{2}Amazon\quad
\textsuperscript{3}SUSTech\quad
\textsuperscript{4}Zhejiang University \\[0.1em]
\textsuperscript{5}Yale University\quad
\textsuperscript{6}Carnegie Mellon University\quad
\textsuperscript{7}HKUST (GZ) \\[0.1em]
\textsuperscript{8}Nanyang Technological University\quad
\textsuperscript{9}Microsoft}
}}

\begin{document}
\maketitle

\begin{abstract}

Large-scale data collection is essential for developing personalized training data, mitigating the shortage of training data, and fine-tuning specialized models. However, creating high-quality datasets quickly and accurately remains a challenge due to annotation errors, the substantial time and costs associated with human labor. To address these issues, we propose Automatic Dataset Construction (ADC), an innovative methodology that automates dataset creation with negligible cost and high efficiency. Taking the image classification task as a starting point, ADC leverages LLMs for the detailed class design and code generation to collect relevant samples via search engines, significantly reducing the need for manual annotation and speeding up the data generation process. To demonstrate ADC at scale, we construct Clothing-ADC: a dataset of over 1 million images spanning 12 main classes and 12,000 fine-grained subclasses. Our automated curation achieves 79\% agreement with human annotators and reduces label noise from 22.2\% to 10.7\%. Despite these advantages, ADC also encounters real-world challenges such as label errors (label noise) and imbalanced data distributions (label bias). We provide open-source software that incorporates existing methods for label error detection, robust learning under noisy and biased data, ensuring a higher-quality training data and more robust model training procedure. Furthermore, we design three benchmark datasets focused on label noise detection, label noise learning, and class-imbalanced learning. These datasets are vital because there are few existing datasets specifically for label noise detection, despite its importance. Finally, we evaluate the performance of existing popular methods on these datasets, thereby facilitating further research in the field.\footnote{The Clothing-ADC dataset is publicly available at \url{https://huggingface.co/datasets/mikelmh025/ClothingADC}.}

\end{abstract}

\section{Introduction}

In the era of Large Language Models (LLMs), the literature has observed an escalating demand for fine-tuning specialized models \citep{benary2023leveraging,porsdam2023autogen,wozniak2024personalized}, highlighting the urgent need for customized datasets \citep{wu2023tidybot,lyu2023llm,tan2024democratizing}. 

\vspace{0.1in}
\noindent\textbf{Traditional Dataset Construction (TDC)} typically involves sample collection followed by labor-intensive annotation, requiring significant human efforts \citep{xiao2015learning, krizhevsky2009learning, wei2021learning, liu2015faceattributes}. Consequently, TDC is often hindered by the limitations of human expertise, leading to suboptimal design \citep{ramaswamy2023overlooked}, data inaccuracies \citep{natarajan2013learning,liu2015classification,li2017webvision,xiao2015learning,wei2022learning}, and extensive manual labor \citep{chang2017revolt,kulesza2014structured}. Furthermore, certain datasets are inherently challenging or risky to collect manually, such as those for fall detection in elderly individuals, dangerous activities like extreme sports, and network intrusion detection. Therefore, there is a growing need for more automated and efficient data collection methods to enhance accuracy and efficiency in dataset creation \citep{bansal2021does,bansal2021most,han2021iterative}. To address these challenges, we propose the \textbf{Automatic Dataset Construction (ADC)}, an innovative approach designed to construct customized large-scale datasets with minimal human involvement. Our methodology reverses the traditional process by starting with detailed annotations that guide sample collection. This significantly reduces the workload, time, and cost associated with human annotation, making the process more efficient and targeted for LLM applications, ultimately outperforming traditional methods. 
\begin{figure*}[!t]
  \vspace{-0.1in}  
  \centering
  \includegraphics[width=1.0\linewidth]{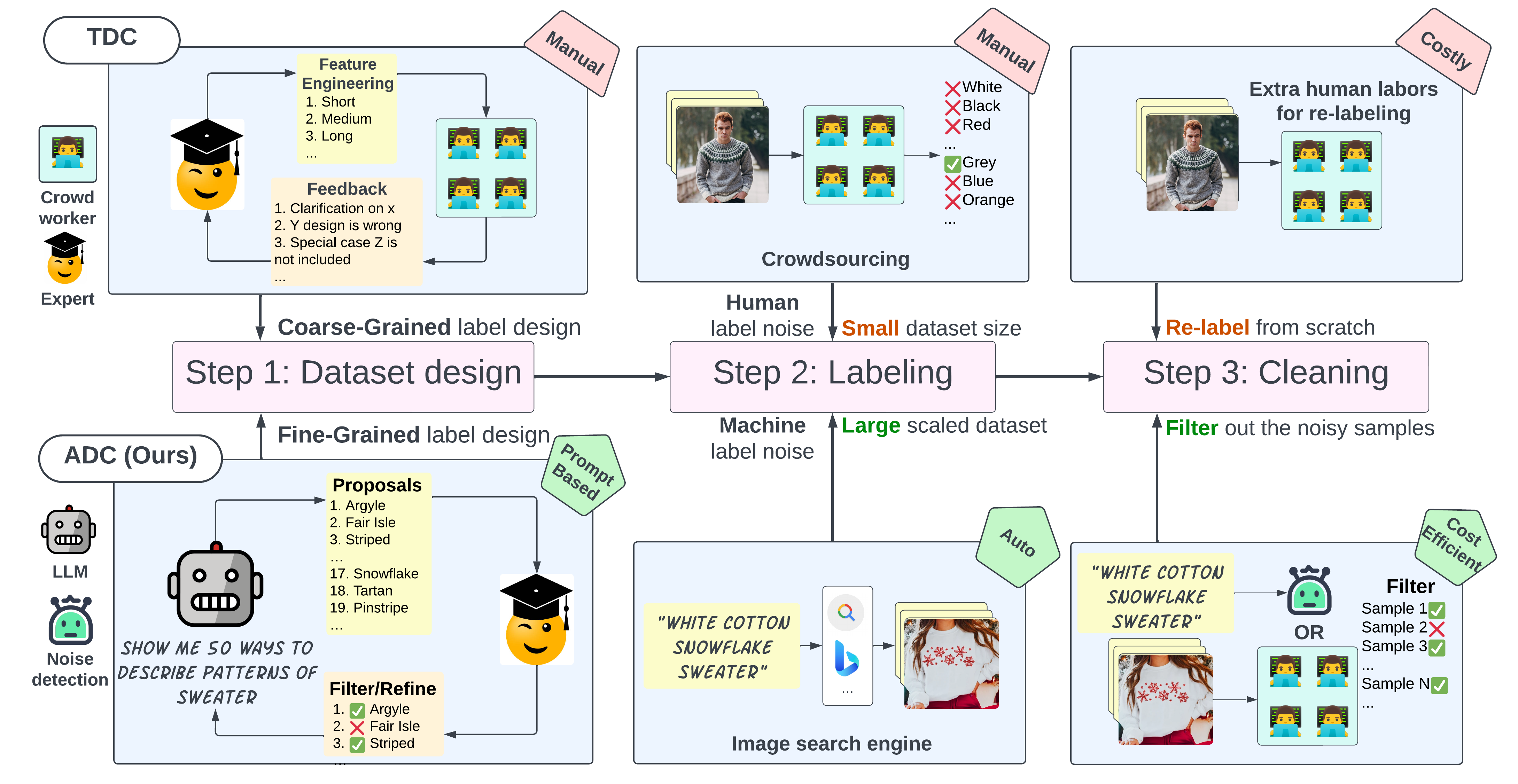}
  \vspace{-0.2in}
  \caption{\textbf{Comparisons of key steps in dataset construction}. In \textbf{Step 1}: Dataset design, ADC utilizes LLMs to search the field and provide instant feedback, unlike traditional methods that rely on manual creation of class names and refine through crowdsourced worker feedback. In \textbf{Step 2}: Labeling, ADC reduces human workload by flipping the data collection process, using targets to search for samples. In \textbf{Step 3}: Cleaning, Instead of re-labeling from scratch, ADC automatically filters out the majority of wrongly samples, by using label noise detection methods. Additional cost efficient human effort can be introduced for a higher quality dataset.}
    \vspace{-0.1in}
  \label{fig:datasetCreation}
\end{figure*}
\vspace{0.1in}
\noindent\textbf{Traditional-Dataset-Construction v.s. Automatic Dataset Construction} Figure \ref{fig:datasetCreation} illustrates the difference between Traditional Dataset Construction (TDC) and Automatic Dataset Construction (ADC). TDC typically unfolds in two stages: developing classification categories and employing human labor for annotation. Creating comprehensive categories requires deep domain knowledge and experience, tasks that even expert researchers find challenging \citep{ramaswamy2023overlooked}. Crowdsourcing is often used to refine these categories, but it increases time and costs without necessarily improving label quality \citep{chang2017revolt, kulesza2014structured}. Annotation by human workers introduces label noise, which impacts dataset reliability, even when multiple inputs are aggregated \citep{sheng2008get}. In contrast, ADC offers improvements at each key step. In the ``Dataset design", ADC uses LLMs to automate field searches and provide instant feedback, unlike traditional manual class and attribute creation. In the sample annotation steps, ADC reverses the labeling process by using predefined targets to search for samples, human annotators are then instructed to filter noisy labeled samples, significantly reducing the need for costly human annotation.

Our main contributions can be summarized as follows:

\squishlist
    \item \textbf{The Automatic-Dataset-Construction (ADC) Pipeline:} We introduce Automatic-Dataset-Construction (ADC), an automatic data collection pipeline that requires minimal human efforts, tailored for specialized large-scale data collection. The pipeline is easily adaptable to any image-related high-quality dataset construction task.
     \item \textbf{Software Efforts for Addressing Dataset Construction Challenges:}
    We explore several challenges observed in real-world dataset construction, including detecting label errors, learning with noisy labels, and class-imbalanced learning. To improve the quality of the constructed data and model training, we provide open-source software that incorporates existing solutions to these challenges.
    \item \textbf{Dataset and Benchmark Efforts:} Leveraging ADC, we developed Clothing-ADC, an image dataset containing one million images with over 1,000 subclasses for each clothing type, publicly available at \href{https://huggingface.co/datasets/mikelmh025/ClothingADC}{HuggingFace}. Our dataset offers a rich hierarchy of categories, creating well-defined sub-populations that support research on a variety of complex and novel tasks. To further facilitate the exploration of the aforementioned challenges (label noise detection and learning, class-imbalanced learning), we customize three benchmark subsets and provide benchmark performances of the implemented methods in our software. This offers researchers a platform for performance comparisons, enhancing the evaluation and refinement of their approaches.
\squishend

\section{Automatic-Dataset-Construction (ADC)} \label{sec:adc}


Traditional methods are invaluable for discovering new knowledge, particularly in fields like citizen science. The efforts of experts in these domains are irreplaceable, and we respect the dedication required to collect and annotate data in these contexts. However, collecting a dataset from the traditional pipeline requires tens of thousand of human labor hours to annotate each sample \citep{van2018inaturalist, deng2009imagenet}. Despite the high effort from human experts, obtaining a clean dataset is very hard under traditional collection methods \citep{northcutt2021pervasive}.

Our proposed ADC pipeline serves a different purpose. Rather than attempting to replace human experts by synthetic labels from models, our ADC provides assistance in collecting existing data from the internet. In this section, we discuss the detailed procedure of ADC, as well as an empirical application.




\subsection{The ADC pipeline}
The ADC pipeline generates datasets with finely-grained class and attribute labels, utilizing data diagnostic software to perform data curation. Below, we provide a step-by-step guide to collecting the Clothing-ADC, a clothes image dataset, along with an overview of its statistics and key information. The overall Automatic-Dataset-Construction (ADC) pipeline is illustrated in Figure \ref{fig:datasetCreation}.

\vspace{0.1in}
\noindent\textbf{Step 1: Dataset design with large language models (LLM)}

\squishlist

    \item \textbf{Detailed Categories Identification:} LLMs assist researchers in conducting a more thorough search in the field by processing and analyzing numerous concepts simultaneously, unlike humans who may overlook certain factors when faced with a large volume of concepts \citep{ramaswamy2023overlooked}. We utilize LLMs to identify attribute types for each class. Then use a prompt of \textit{"Show me <30-80> ways to describe <Attribute> of <Class>"} to generate the proposed subclasses. To quantify the LLM's added value, we compare LLM-generated categories to those produced by manual brainstorming: manual enumeration yields roughly 10 color options, 7 materials, and 6 patterns for clothing, whereas GPT-4 generates 30--50 options per attribute. Critically, the LLM surfaces domain-specific vocabulary that manual efforts routinely miss---for colors, technical dyeing terms; for materials, luxury and technical fabrics such as Gore-Tex, Modal, Alpaca, and Cashmere; and for patterns, specialized textile designs such as Fair Isle and Houndstooth. These additions substantially enrich the diversity of collected samples.

\begin{figure}[!htb]
\vspace{-0.1in}
  \centering
  \includegraphics[width=0.96\linewidth]{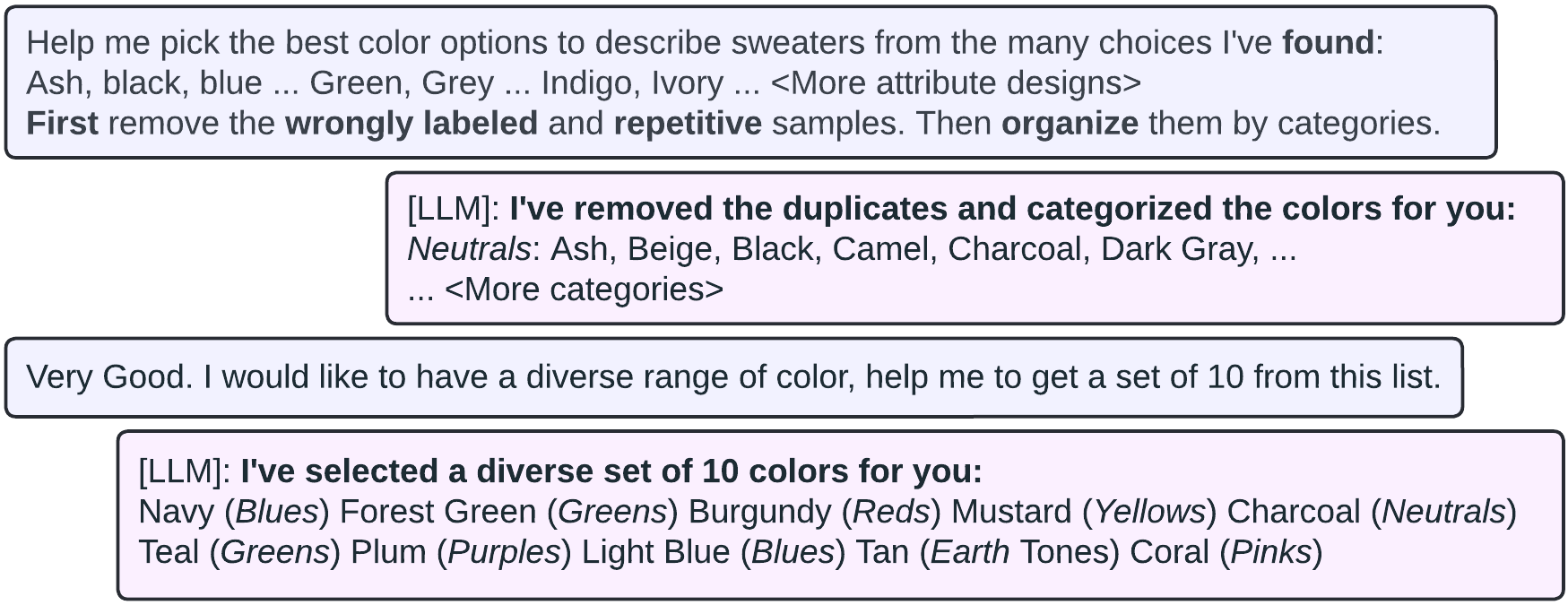}
\vspace{-0.05in}
  \caption{Example of using LLM for iterative refinement of attribute designs.}
\vspace{-0.2in}

  \label{fig:datasetIterative}
\end{figure}

    \item \textbf{Iterative Refinement:} The initial category list generated by the LLM undergoes review and refinement either by domain experts or through self-examination by the LLM itself, ensuring alignment with specific application or research needs, as shown in Figure \ref{fig:datasetIterative}. This iterative refinement process enables the creation of a high-quality dataset with finely-grained class labels. Additionally, this approach facilitates rapid iterative feedback during the design phase, offering a significant advantage over traditional methods that rely on annotator feedback during the test run annotation phase. This acceleration enables researchers to explore and refine their ideas more efficiently, resulting in better dataset quality and reduced development time.

    \item \textbf{LLM Hallucination Issues:} LLMs and VLMs are capable of providing synthetic sample labeling. However, they tend to hallucinate excessively and uncontrollably \citep{xu2024hallucination,huang2023survey}.  To create a responsible dataset, we limit the use of LLMs to the dataset design phase, where they assist human designers. Any hallucinated or inaccurate labels should be caught.
\squishend
\vspace{0.1in}
\noindent\textbf{Step 2: Automated Labeling}
For image data collection and labeling, ADC utilizes APIs provided by Google or Bing for automated querying, guaranteeing real samples are collected from the web. Each category and attribute identified in the first step can be used to formulate search queries, which is the sample label also. 

\vspace{0.1in}
\noindent\textbf{Step 3: Data Curation and Cleaning}

\squishlist

\item \textbf{Duplicate Image Removal:} Web-based collection inevitably introduces duplicate images, both within a single search query and across different queries that share overlapping results. We apply perceptual hashing (pHash) and flag pairs with a Hamming distance below 3 as near-duplicates. Within-query duplicates are rare (0.38\%), reflecting the search engine's internal deduplication. Cross-query duplicates are more prevalent: before cleaning, 64.16\% of images appeared in more than one query; after removing duplicates, this rate drops to 52.33\%, yielding a substantially more diverse training set.

\item \textbf{Algorithmic Label Noise Detection:} For applications where some label noise can be tolerated, existing data curation software capable of identifying and filtering out irrelevant images, such as Docta, CleanLab , and Snorkel \footnote{Docta:\url{www.docta.ai}, CleanLab:\url{www.cleanlab.ai}, Snorkel:\url{www.snorkel.ai}}, etc. For example, these tools can identify when an item is mislabeled regarding its type, material, or color. Finally, ADC aggregates the suggested labels recommended by the dataset curation software and removes potentially mislabeled or uncertain samples. For illustration, we adopt a data-centric label curation software (Docta) in Algorithm \ref{alg:docta}. 
The high-level idea of this algorithm is to estimate the essential label noise transition matrix $T\_Est$ without using ground truth labels, achieved through the consensus equations (\textbf{Part A}). Following this, Algorithm \ref{alg:docta} identifies those corrupted instances via the cosine similarity ranking score among features as well as a well-tailored threshold based on the obtained information (i.e., $T\_Est$), and then relabels these instances using KNN-based methods (\textbf{Part B}). For more details, please refer to work \citep{zhu2023unmasking, zhu2021clusterability,zhu2022detecting}.




\begin{figure}[t]
\vspace{-0.1in}
    \begin{algorithm}[H]
    \caption{Data centric curation (Docta)}\label{alg:docta}
    \begin{algorithmic}[1]
    {\small
    \Procedure{Docta}{noisyDataset, preTrainedModel}
    \State \textbf{Part A: Feature Extraction and Noise Analysis}
    \State \hspace{1.0em} $features \leftarrow$ EncodeImages(noisyDataset, preTrainedModel)
    \State \hspace{1.0em} $T_Est \leftarrow$ EstimateTransitionMatrix($features, noisyLabels$)
    \State \textbf{Part B: Label Correction}
    \State \hspace{1.0em} $corrupted\_Instances$ 
    \State \hspace{2.5em} $\leftarrow$ SimiFeat-rank($features, noisyLabels, T\_Est$)
    \State \hspace{1.0em} $cured\_Labels$ 
    \State \hspace{2.5em} $\leftarrow$ KNN-based Relabeling($corrupted\_Instances$)
    \State \textbf{Return} $cured\_Labels$
    \EndProcedure}
    \end{algorithmic}
    \end{algorithm}
\vspace{-0.3in}
\end{figure}

\squishend

\squishlist
\item \textbf{Cost Efficient Human-in-the-Loop:} For domains requiring clean data, we advocate for human involvement in addition to algorithmic approaches to ensure perfect annotations. Unlike traditional pipelines where humans are asked to relabel samples from scratch, our ADC pipeline provides a large amount of noisy labeled samples for humans to review and select the accurate ones. This approach is mentally easier and results in a clean dataset, as the selected samples have guaranteed human and machine label agreements. Analyses of human votes are in Appendix \ref{app:statistics}.
\squishend



\subsection{Clothing-ADC}
\begin{figure*}[!htb]
    \centering
    {\begin{minipage}[c]{0.31\textwidth}  
        \centering

        \resizebox{\textwidth}{!}{        
        \begin{tabular}{|l|r|}
            \hline \rowcolor{blue!10}\multicolumn{2}{|c|}{ \textbf{Dataset Overview} } \\
            \hline \hline Number of Samples & $1,076,738$ \\
            \hline Resolution & $256 \times 256$ \\
            \hline \hline \rowcolor{blue!10}\multicolumn{2}{|c|}{ \textbf{Dataset Split} } \\
            \hline Train set(with web noise) & $1,036,738$ \\
            \hline Evaluation set (Clean) & 20,000 \\
            \hline Test set (Clean) & 20,000 \\
            \hline \hline \rowcolor{blue!10}\multicolumn{2}{|c|}{ \textbf{Classification Structure} } \\
            \hline Main Class & 12 \\
            \hline Total Subclasses & 12,000 \\
            \hline \hline \rowcolor{blue!10}\multicolumn{2}{|c|}{ \textbf{Subclass Details} } \\
            \hline Attribute (Color) & 10 \\
            \hline Attribute (Material) & 10 \\
            \hline Attribute (Pattern) & 10 \\
            \hline Ave. Samples per attribute & 89.73 \\
            \hline
        \end{tabular}
        }
        \captionof{table}{Dataset information summary of Clothing-ADC Dataset.}
        \label{table:clothingadc_stat}
    \end{minipage}%
    \begin{minipage}[c]{0.68\textwidth}  
        \vspace{-0.25in}
        \centering
        \includegraphics[width=\textwidth]{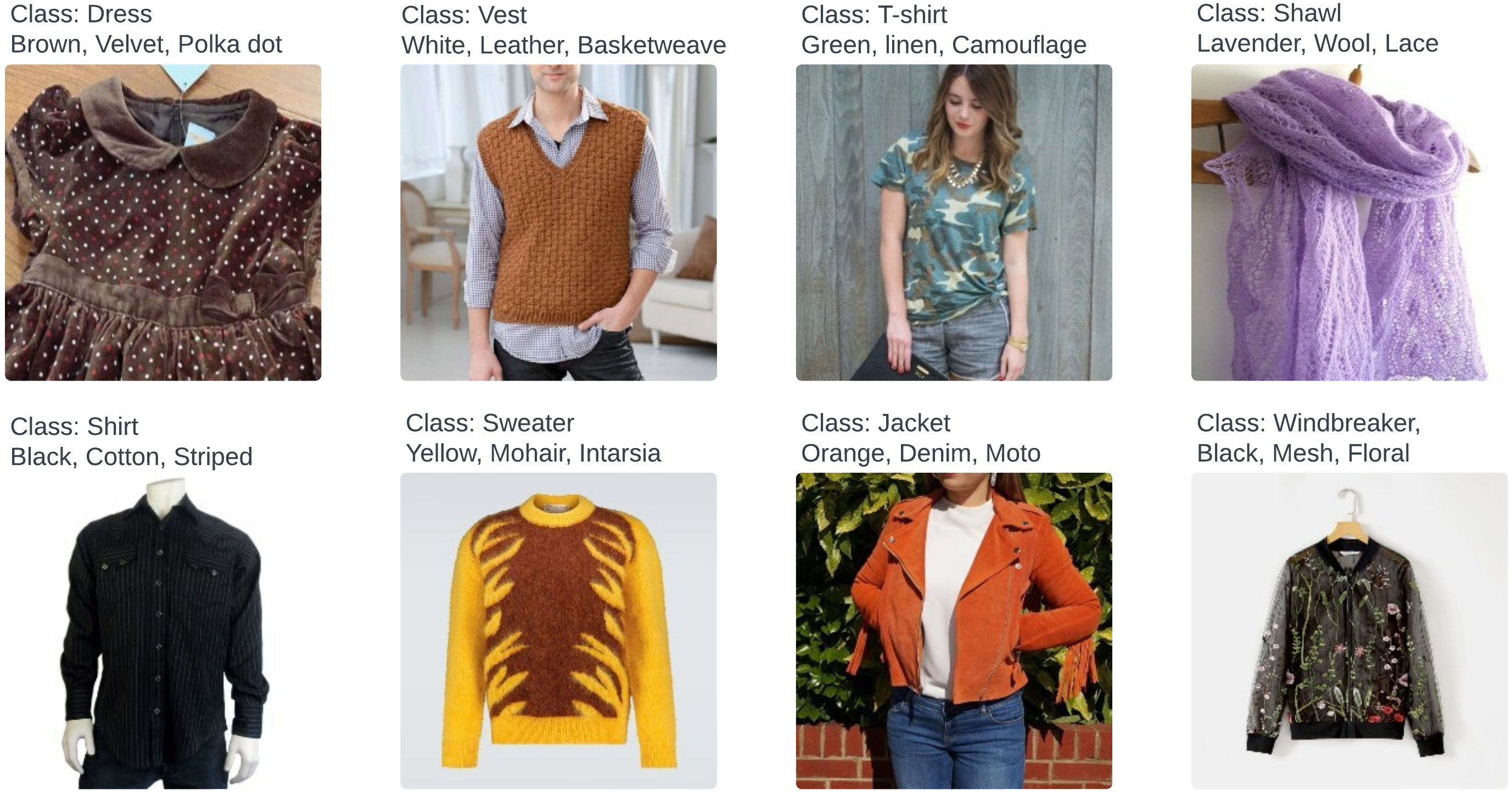}
        \caption{Samples from the collected Clothing-ADC Dataset}
        \label{fig:clothingadc_sample}
    \end{minipage}}
    \vspace{-0.2in}
\end{figure*}

To illustrate the ADC pipeline, we present the Clothing-ADC dataset, which comprises a substantial collection of clothing images. The dataset is publicly available at \href{https://huggingface.co/datasets/mikelmh025/ClothingADC}{\textbf{HuggingFace}}. The dataset includes 1,076,738 samples, with 20,000 allocated for evaluation, another 20,000 for testing, and the remaining samples used for training. Each image is provided at a resolution of 256x256 pixels. The dataset is categorized into 12 primary classes, encompassing a total of 12,000 subclasses, with an average of 89.73 samples per subclass. Detailed statistics of the dataset are provided in Table \ref{table:clothingadc_stat}. The following subsection elaborates on the dataset construction process in comprehensive detail. Other ADC application examples are in Appendix \ref{app:demo_applicaiton}. 

\vspace{0.1in}
\noindent\textbf{Subclass Design} Utilizing GPT-4, we identified numerous attribute options for each clothing type. For example, in the case of sweaters, we recognized eight distinct attributes: color, material, pattern, texture, length, neckline, sleeve length, and fit type. The language model was able to find 30-50 options under each attribute. Our Clothing-ADC dataset includes the three most common attributes: color, material, and pattern, with each attribute having ten selected options. This results in 1000 unique subclasses per clothing type. The selected attributes are detailed in Table \ref{Tab:dataset_attributes_full} (Appendix).

\vspace{0.1in}
\noindent\textbf{Data Collection} The ADC pipeline uses the Google Image API to collect clothing images by formulating queries that include attributes such as "Color + Material + Pattern + Cloth Type" (e.g., "white cotton fisherman sweater"). Figure \ref{fig:clothingadc_sample} shows examples of these queries and the corresponding images retrieved. The relevance of the search results tends to decline after a significant number of samples are gathered, leading us to set a cutoff threshold of 100 samples per query. After removing broken links and improperly formatted images, each subclass retained roughly 90 samples. These queries generated noisy, webly-labeled data for the training set.

\vspace{0.1in}
\noindent\textbf{Automatic label noise curation} The collected samples may suffer from web-based label noise, where annotations might be incorrect due to mismatches provided by search engines, the traditional approach typically involves manually re-labeling existing annotations and aggregating multiple human votes per label to achieve a higher-quality dataset. To address this issue, ADC uses label noise detection methods to filter out the majority of wrongly labeled samples. To validate effectiveness of the autocuration, we conducted a human-in-the-loop evaluation, comparing automated curation results against human curation. This analysis revealed a \textbf{79.0\%} agreement rate between automated and human curation decisions. We observed a reduction in label noise from \textbf{22.2\%} to \textbf{10.7\% }after applying the ADC auto-curation, demonstrating its effectiveness in improving dataset quality. In Section \ref{sec:label_noise_detection}, we show a detailed discussion and benchmark of label noise detection methods.


\vspace{0.1in}
\noindent\textbf{Creating Test Set} The automatic collection and curation pipeline results in a large-scale data set with fine-grained labeling. Despite the fact that majority of the label noise has been removed, the test set requires clean labels. Thus, ADC introduces additional human effort achieving the 100\% human-machine agreement rate. 

Instead of manually relabeling all samples like traditional methods, ADC pipeline enhances human cost efficiency by presenting annotators with a set of samples that share the same machine-generated label. Annotators are then tasked with selecting a subset of correctly labeled samples, choosing a minimum of four samples out of twenty. This method significantly reduces both manual effort and difficulty, encouraging annotators to critically evaluate machine-generated labels and thereby reducing the effect of human over-trust in AI answers \citep{bansal2019updates,bansal2021does}. The samples selected through this process are considered ``clean" labels, representing a consensus between human judgment and machine-generated labels~\citep{liu2023humans}. We list a detailed analysis of human cost against traditional methods in Appendix \ref{sec:appendix human curation cost analysis}.

\vspace{0.1in}
\noindent\textbf{Compare With Existing Datasets} Table \ref{tab:dataset_summary} provides an insightful comparison between existing datasets and Clothing-ADC. Briefly speaking, compared with existing datasets, the ADC pipeline is able to help humans without domain expertise to create fine-grained attributes for the dataset, and automatic annotation and label cleaning drastically eliminate human effort during label creation. The fine-grained design of attributes reduces subclass imbalance issue in the collection dataset.

\begin{table*}[!htb]
\centering
\caption{Our ADC pipeline creates a large-scale image classification dataset with a clean test set. Most existing datasets require human effort for labeling, whereas our pipeline can automatically annotate and clean the data. While Clothing-ADC provides fine-grained attribute labels, our dataset design does not require human expertise in the field.}
\resizebox{.88\textwidth}{!}{
    \begin{tabular}{|c|c|c|c|c|c|c|}
    \hline
    \cellcolor{blue!10} \textbf{Dataset} & \cellcolor{blue!10} \textbf{\# Train/Test} & \cellcolor{blue!10} \textbf{\# Classes} & \cellcolor{blue!10} \textbf{Noise Rate(\%)} & \cellcolor{blue!10} \textbf{Has Attributes} & \cellcolor{blue!10} \textbf{Auto annotation} & \cellcolor{blue!10} \textbf{Require expert?} \\ \hline \hline
    \textbf{iNaturalist} \citep{van2018inaturalist} & 579k/279k     & 54k    & Close to 0        & \xmark & \xmark &  \cmark \\ \hline
    \textbf{WebVision} \citep{li2017webvision}      & 2.4M/100k     & 1000   & 20                & \xmark & \cmark &  \cmark \\ \hline
    \textbf{ANIMAL-10N} \citep{song2019selfie}      & 50k/10k       & 10     & 8                 & \xmark & \xmark &  \xmark \\ \hline
    \textbf{CIFAR-10N} \citep{wei2021learning}      & 50k/10k       & 10     & 9.03/25.60/40.21$^*$  & \xmark & \xmark &  \xmark \\ \hline
    \textbf{CIFAR-100N} \citep{wei2021learning}     & 50k/10k       & 100    & 25.6/40.2$^*$         & \xmark & \xmark &  \xmark \\ \hline
    \textbf{Food-101N} \citep{bossard14}            & 75.75k/25.25k & 101    & 18.4              & \xmark & \xmark &  \cmark \\ \hline
    \textbf{Clothing1M} \citep{xiao2015learning}    & 1M in all     & 14     & 38.5              & \xmark & \xmark &  \cmark \\ \hline
    \textbf{Clothing-ADC (Ours})                   & 1M/20k        & 12     & 22.2-32.7       & 12k    & \cmark &  \xmark \\ \hline
    \end{tabular}
}
\par\noindent{\small $^*$Multiple values correspond to different human annotation settings reported in the original paper (clean / worst-case / aggregate noise rates).}
\label{tab:dataset_summary}
\end{table*}

\section{Challenge one: dealing with imperfect data annotations}\label{sec:cha_noise}

The first pervasive and critical challenge during the automatic dataset construction lies in the prevalence of noisy/imperfect labels. This issue is intrinsic to web-sourced data, which, although rich in diversity, often suffers from inaccuracies due to the uncurated nature of the internet. These errors manifest as mislabeled images, inconsistent tagging, and misclassified attributes, introducing non-negligible noise into the dataset that may adversely affect the training and performance of machine learning models. The following discussion bridges the gap between imperfect data and curated data via mining and learning with label noise, to refine data quality, enhance label accuracy, and ensure the reliability of Auto-Dataset-Construction (ADC) for high-stakes AI applications.

\vspace{0.1in}
\noindent\textbf{Formulation}
Let $D := \{(x_n, y_n)\}_{n \in [N]}$ represent the training samples for a $K$-class classification task, where $[N] := \{1, 2, ..., N\}$. Suppose that these samples $\{(x_n, y_n)\}_{n \in [N]}$ are outcomes of the random variables $(X, Y) \in \mathcal{X} \times \mathcal{Y}$, drawn from the joint distribution $\mathcal{D}$. Here, $\mathcal{X}$ and $\mathcal{Y}$ denote the spaces of features and labels, respectively. However, classifiers typically access a noisily labeled training set $\widetilde{D} := \{(x_n, \tilde{y}_n)\}_{n \in [N]}$, assumed to arise from random variables $(X, \widetilde{Y}) \in \mathcal{X} \times \widetilde{\mathcal{Y}}$, drawn from the distribution $\widetilde{\mathcal{D}}$. It is common to observe instances where $y_n \neq \tilde{y}_n$ for some $n \in [N]$. The transition from clean to noisy labels is typically characterized by a noise transition matrix $T(X)$, defined as $T_{i,j}(X) := \mathbb{P}(\widetilde{Y} = j \mid Y = i, X)$ for all $i, j \in [K]$ \citep{natarajan2013learning, liu2015classification, patrini2017making}.


\subsection{The challenge of label noise detection}
\label{sec:label_noise_detection}


While employing human annotators to clean data is effective in improving label quality, it is often prohibitively expensive and time-consuming for large datasets. A practical alternative is to enhance label accuracy automatically by first deploying algorithms to detect potential errors within the dataset and then correcting these errors through additional algorithmic processing or crowdsourcing.

\vspace{0.2in}
\subsubsection{Existing approaches to detect label noise}
\paragraph{\textbf{Learning-Centric Approaches:}} Learning-centric approaches often leverage the behavior of models during training to infer the presence of label errors based on how data is learned. One effective strategy is confidence-based screening, where labels of training instances are scrutinized if the model's prediction confidence falls below a certain threshold. This approach assumes that instances with low confidence scores in the late training stage are likely mislabeled \citep{northcutt2021confident}. Another innovative technique involves analyzing the gradients of the training loss w.r.t. input data. \cite{pruthi2020estimating} utilize gradient information to detect anomalies in label assignments, particularly focusing on instances where the gradient direction deviates significantly from the majority of instances. Researchers have also utilized the memorization effect of deep neural networks, where models tend to learn clean data first and only memorize noisy labels in the later stages of training. Techniques that track how quickly instances are learned during training can thus identify noisy labels by focusing on those learned last \citep{han2019deep,liu2020early,xia2020robust}. 

\paragraph{\textbf{Data-Centric Approaches:}}
Data-centric methods focus on analyzing data features and relationships rather than model behavior for detection. The ranking-based detection method \citep{brodley1999identifying} ranks instances by the likelihood of label errors based on their alignment with model predictions. An ensemble of classifiers evaluates each instance, flagging those that consistently deviate from the majority vote as noisy. Neighborhood Cleaning Rule \cite{laurikkala2001improving} uses the $k$-nearest neighbors algorithm to check label consistency with neighbors, identifying instances whose labels conflict with the majority of their neighbors as potentially noisy. \cite{zhu2022detecting} propose advanced data-centric strategies for detecting label noise without training models. Their local voting method uses neighbor consensus to validate label accuracy, effectively identifying errors based on agreement within the local feature space. 
\subsubsection{Clothing-ADC in label noise detection} 

\paragraph{\textbf{Setup:}} We prepared a subset of 20,000 samples from the Clothing-ADC dataset for the label noise detection task, including both noisy and clean labels. We collected three human annotations for each image via Amazon MTurk. Annotators were instructed to classify the labels as correct, unsure, or incorrect. Each sample received three votes. Based on these annotations, we determined the noise rate to be 22.2\%-32.7\%. Using majority vote aggregation implies uncertainty of the label correctness. By using a more stringent aggregation criterion, more samples are considered as noisy labeled. Under the extreme case where any doubts from any human annotator can disqualify a sample, our auto collected dataset still retains 61.3\% of its samples. For a detailed distribution of human votes, see Table \ref{tab:label_noise_human_votes} in the Appendix. 






\paragraph{\textbf{Benchmark Efforts}} 

\begin{table}[ht]
\centering
\caption{$F_1$-Score comparisons among several label noise detection methods on Clothing-ADC.}

\resizebox{\textwidth}{!}{%
\begin{tabular}{c|cccc}

\hline
\cellcolor{blue!10}\textbf{Methods}  & \textbf{CORES} \cite{cheng2020learning} & \textbf{CL} \cite{northcutt2021confident} & Deep\textbf{ $k$-NN} \cite{papernot2018deep}  & \textbf{Simi-Feat} \cite{zhu2022detecting}                                           \\\hline



\cellcolor{blue!10}$F_1$-\textbf{Score} & 0.4793 & 0.4352&0.3991 & 0.5721                             \\\hline
\end{tabular}}

\label{tab:label_noise_detection}
\end{table}

Detection performance comparisons of certain existing solutions are given in Table \ref{tab:label_noise_detection}. We adopt ResNet-50 \citep{he2016deep} as the backbone model to extract the feature here. For each method, we use the default hyper-parameter reported in the original papers. All methods are tested on 20,000 points and predict whether the data point is corrupted or not. We follow \cite{zhu2022detecting} to apply the baseline methods to our scenario. In Table \ref{tab:label_noise_detection}, the performance is measured by the $F_1$-score of the detected corrupted instances, which is the harmonic mean of the precision and recall, i.e., 
$F_1=\frac{2}{\text{Precision}^{-1}+\text{Recall}^{-1}}.$
Let $v_n=1$ indicate that the $n$-th label is detected as a noisy/wrong label, and $v_n=0$ otherwise. Then, the precision and recall of detecting noisy labels can be calculated as:
$\text{Precision}=\frac{\sum_n \mathbbm{1}(v_n=1, \tilde{y}_n\neq y_n)}{\sum_n \mathbbm{1}(v_n=1)},\quad  \text{Recall}=\frac{\sum_n \mathbbm{1}(v_n=1, \tilde{y}_n\neq y_n)}{\sum_n \mathbbm{1}(\tilde{y}_n\neq y_n)}.$ 

\vspace{0.05in}

Simi-Feat\cite{zhu2022detecting} has the best performance in the benchmark. We observed a reduction in label noise from \textbf{22.2\%} to \textbf{10.7\% } in Clothing-ADC, demonstrating its effectiveness in improving dataset quality.


\subsection{The challenge of learning with noisy labels}
\label{sec:lablel_noise_learning}

Another technique is robust learning that can effectively learn from noisy datasets without being misled by incorrect labels, thus maintaining high accuracy and reliability in real-world applications.

\subsubsection{Existing approaches to learn with label noise} In this subsection, we contribute to the literature with robust learning software; all covered methods can be mainly summarized into the following three categories: robust loss functions, robust regularization techniques, and multi-network strategies.

\vspace{0.1in}
\noindent\textbf{Robust Loss Designs:} Loss Correction modifies the traditional loss function to address label noise by incorporating an estimated noise transition matrix, thereby recalibrating the model's training focus \citep{patrini2017making}. Loss-Weighting strategies mitigate the impact of noisy labels by assigning lower weights to likely mislabeled instances, reducing their influence on the learning process \citep{liu2015classification,ren2018learning}. Symmetric Cross-Entropy Loss balances the contributions of correctly labeled and mislabeled instances, improving the model's resilience to label discrepancies \citep{wang2019symmetric}. Generalized Cross-Entropy Loss, derived from mean absolute error, offers enhanced robustness against outliers and label noise \citep{zhang2018generalized}. Peer Loss Functions form a family of robust loss functions \citep{liu2020peer,wei2020optimizing,cheng2020learning}, leveraging predictions from peer samples as regularization to adjust the loss computation, thereby increasing resistance to noise.

\vspace{0.1in}
\noindent\textbf{Robust Regularization Techniques:}
Regularization techniques are designed to constrain or modify the learning process, thereby reducing the model's sensitivity to label noise. Mixup \citep{zhang2017mixup} generates synthetic training examples by linearly interpolating between pairs of samples and their labels, enhancing model generalization and smoothing label predictions. Label Smoothing \citep{muller2019does, lukasik2020does} combats overconfidence in unreliable labels by adjusting them towards a uniform distribution. Negative Label Smoothing \citep{wei2022smooth} refines this approach by specifically adjusting the smoothing process for negative labels, preserving model confidence in high-noise environments. Early-Learning Regularization tackles the issue of early memorization of noisy labels by dynamically adjusting regularization techniques during the initial training phase \citep{liu2020early,xia2020robust}.

\vspace{0.1in}
\noindent\textbf{Multi-Network Strategies:} Employing multiple networks can enhance error detection and correction through mutual agreement and ensemble techniques. In Co-teaching, two networks concurrently train and selectively share clean data points with each other, mitigating the memorization of noisy labels \citep{han2018co}. MentorNet \citep{jiang2018mentornet} equips a student network with a curriculum that emphasizes samples likely to be clean, as decided by the observed dynamics of a mentor network. DivideMix leverages two networks to segregate the data into clean and noisy subsets with a mixture model, allowing for targeted training on each set to manage label noise \citep{Li2020DivideMix}.

\subsubsection{Clothing-ADC in label noise learning}

\begin{table}[ht]
\centering
\caption{Label noise learning results on Clothing-ADC and Clothing-ADC (tiny), reporting test accuracy on the held-out clean set. Clothing-ADC (tiny) reports mean$\pm$std over 3 seeds; non-overlapping intervals indicate significance (e.g., TaylorCE $71.11{\pm}0.07$ vs.\ CE $67.72{\pm}0.40$, gap $>8\times$ std).}

\resizebox{.68\textwidth}{!}{%
\centering
\begin{tabular}{|ccc|}
\hline
\multicolumn{1}{|c|}{\cellcolor{blue!10}\textbf{Methods / Dataset}}         & \multicolumn{1}{c|}{\cellcolor{blue!10}\begin{tabular}[c]{@{}l@{}}\textbf{Clothing-ADC}\end{tabular}} & \cellcolor{blue!10}\begin{tabular}[c]{@{}l@{}}\textbf{Clothing-ADC (tiny)}\end{tabular} \\ \hline \hline
\multicolumn{1}{|c|}{\textbf{Cross-Entropy}} & \multicolumn{1}{|c|}{74.76}   &  $67.72 \pm 0.40$  \\ 
\multicolumn{1}{|c|}{\textbf{Backward Correction} \citep{patrini2017making}} & \multicolumn{1}{c|}{77.51} & $70.49 \pm 0.06$ \\
\multicolumn{1}{|c|}{\textbf{Forward Correction} \citep{patrini2017making}}  & \multicolumn{1}{c|}{78.45} & $70.60 \pm 0.14$ \\
\multicolumn{1}{|c|}{\textbf{(Positive) LS} \citep{lukasik2020does}} &  \multicolumn{1}{c|}{81.94}   &  $70.67 \pm 0.15$   \\
\multicolumn{1}{|c|}{\textbf{(Negative)} LS \citep{wei2022smooth}} & \multicolumn{1}{c|}{78.65}   &  $70.14 \pm 0.13$   \\
\multicolumn{1}{|c|}{\textbf{Peer Loss} \citep{liu2020peer}} & \multicolumn{1}{c|}{78.58}  &  $	70.92 \pm 0.17$ \\
\multicolumn{1}{|c|}{\textbf{$f$-Div} \citep{wei2020optimizing}} & \multicolumn{1}{c|}{77.43}  & $68.98 \pm 0.22$ \\
\multicolumn{1}{|c|}{\textbf{Divide-Mix} \citep{Li2020DivideMix}} & \multicolumn{1}{c|}{77.00} &      $71.58 \pm 0.11$                                                  \\
\multicolumn{1}{|c|}{\textbf{Jocor} \citep{wei2020combating}} & \multicolumn{1}{c|}{78.47} & $72.81 \pm 0.02$ \\
\multicolumn{1}{|c|}{\textbf{Co-Teaching} \citep{han2018co}} & \multicolumn{1}{c|}{80.49} & $70.55 \pm 0.08$   \\
\multicolumn{1}{|c|}{\textbf{LogitCLIP} \citep{wei2023mitigating}} & \multicolumn{1}{c|}{77.85} & $	70.16 \pm 0.14$ \\
\multicolumn{1}{|c|}{\textbf{TaylorCE} \citep{chen2022improved}} & \multicolumn{1}{c|}{81.87}  & $71.11 \pm 0.07$ \\
\hline
\end{tabular}}

\label{tab:label_noise_learning}
\end{table}

We provide two versions of the Label Noise Learning task, Clothing-ADC and Clothing-ADC (tiny). Specifically, Clothing-ADC leverages the whole available (noisy) training samples to construct the label noise learning task. The objective is to perform class prediction w.r.t. 12 clothes types: 
Sweater, Windbreaker, T-shirt, Shirt, Knitwear, Hoodie, Jacket, Suit, Shawl, Dress, Vest, Underwear. We also provide a tiny version of Clothing-ADC, which contains 50K training images, sharing similar size with certain widely-used ones, i.e., MNIST, Fashion-MNIST, CIFAR-10, CIFAR-100, etc. 

\vspace{0.1in}
\noindent\textbf{Estimated Noise Level of Clothing-ADC:} We selected a subset of 20,000 training samples and asked human annotators to evaluate the correctness of the auto-annotated dataset. After aggregating three votes from annotators, we estimate the noise rate to be 22.2\%-32.7\%, which consists of 10.5\% of the samples having ambiguity and 22.2\% being wrongly labeled. The remaining 77.8\% of the samples were correctly labeled. The detailed distribution of human votes is given in Appendix Table \ref{tab:label_noise_human_votes}. 



\vspace{0.1in}
\noindent\textbf{Benchmark Efforts:}
In this task, we aim to provide the performance comparison among various learning-with-noisy-label solutions. All methods utilize ResNet-50 as the backbone model and are trained for 20 epochs to ensure a fair comparison. We report the model prediction accuracy on the held-out clean labeled test set. For the tiny version, we conduct three individual experiments using three different random seeds and calculate the mean and standard deviation. As shown in Table \ref{tab:label_noise_learning}, certain methods, such as Positive LS and Taylor CE, significantly outperform Cross-Entropy. These results underscore the importance and necessity of pairing ADC with robust learning software. 


\vspace{0.1in}
\noindent\textbf{Benchmark Analysis:} With the increment of dataset size, there is an obvious improvement for the performance. All the noisy learning methods show their effectiveness over the baseline method (Cross-Entropy). TaylorCE achieves the highest performance on both settings, showing the effectiveness of the differential elements. Divide-Mix is also a competitive method because of its dual model architecture. The forward and backward correction is simple but effective. 


\section{Challenge two: dealing with imbalanced data distribution}

\begin{figure*}[!t]
  \centering
  \includegraphics[width=.999\linewidth]{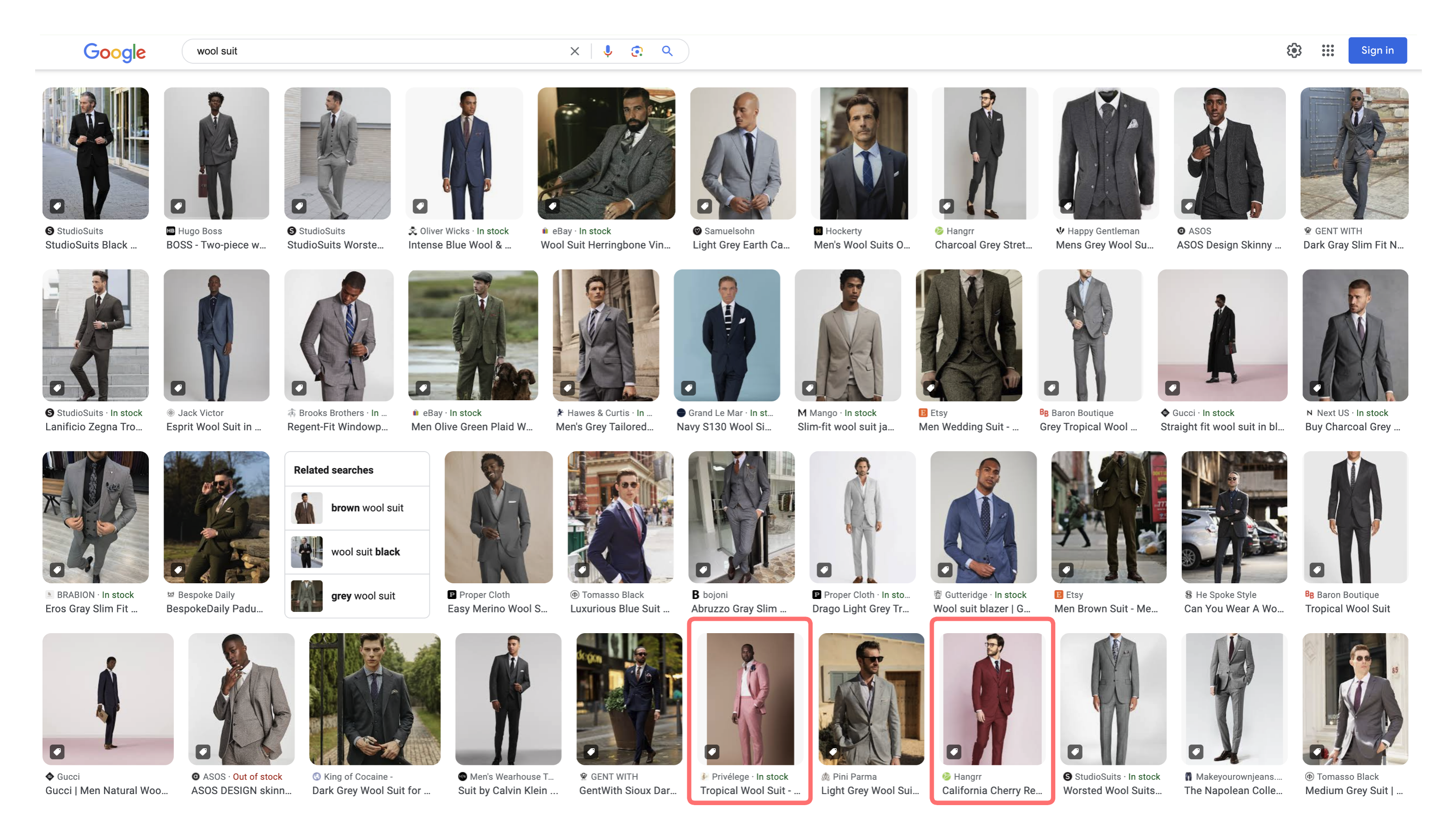}
  \vspace{-0.25in}
  \caption{Long-tailed data distribution is a prevalent issue in many datasets. Searching ``wool suit" in Google image results in dark wool suits, while only a few are of a light color (red/pink).}
  \label{fig:insight}
\end{figure*}

We now discuss another real-world challenge: when imperfect annotations meet with imbalanced class/attribute distributions. As shown in Figure \ref{fig:insight}, long-tailed data distribution is a prevalent issue in web-based datasets: to collect a dataset of wool suits without a specified target color on Google Image, the majority would likely be dark or muted shades (grey, black, navy), with few samples in brighter colors like pink or purple. This natural disparity results in most data points belonging to a few dominant categories, while the remaining are spread across several minority groups.

We are interested in how class-imbalance intervenes with learning. In real-world scenarios, the distribution of classes tends to form a long-tail form, in other words, the head class and the tail class differ significantly in their sample sizes, i.e., $\max_k \mathbb{P}(Y=k) \gg \min_{k'} \mathbb{P}(Y=k')$. 

\subsection{Existing approaches for class imbalance learning}
\vspace{0.1in}
\noindent\textbf{Data-Level Methods}
Data-level methods modify training data to balance class distribution, focusing on adjusting the dataset by increasing minority class instances or decreasing majority class instances. Oversampling increases the number of minority class instances to match or approach the majority class. This can be done through simple duplication \citep{jo2004class} (e.g., random oversampling) or generating synthetic data \citep{chawla2002smote,han2005borderline,bunkhumpornpat2009safe,he2008adasyn}.  Undersampling reduces the number of majority class instances, helping to balance class distributions but potentially discarding useful information \citep{mani2003knn,kubat1997addressing,tomek1976two}.

\vspace{0.1in}
\noindent\textbf{Algorithm-Level Methods}
These methods adjust the training process or model to handle unequal class distributions better. Specifically, cost-sensitive learning assigns different costs to misclassifications of different classes, imposing higher penalties for errors on the minority class \citep{elkan2001foundations}. It modifies the loss function to incorporate misclassification costs, encouraging the model to focus more on minority class errors \citep{kukar1998cost,zhou2005training}. Thresholding adjusts the decision threshold for class probabilities to account for class imbalance. Instead of using a default threshold, different thresholds are applied based on class distribution, modifying the decision process for predicting class labels  \citep{lawrence2002neural, richard1991neural}.


\subsection{Clothing-ADC in class-imbalanced learning}


\begin{table*}[!htb]
    \caption{$\delta$-worst accuracy of class-imbalanced learning baselines on Clothing-ADC CLT dataset.}
    \centering
    \resizebox{\textwidth}{!}{
    \begin{tabular}{l|ccccccccc}
        \toprule
        \rowcolor{blue!10}\textbf{Method} & \multicolumn{3}{c}{\textbf{$\delta=0$ Worst Accuracy}} & \multicolumn{3}{c}{\textbf{$\delta=1$ Worst Accuracy}} & \multicolumn{3}{c}{\textbf{$\delta=\infty$ Worst Accuracy}} \\
        \cmidrule(lr){2-4} \cmidrule(lr){5-7} \cmidrule(lr){8-10}
        & $\rho=10$ & $\rho=50$ & $\rho=100$ & $\rho=10$ & $\rho=50$ & $\rho=100$ & $\rho=10$ & $\rho=50$ & $\rho=100$ \\
        \midrule
        \textbf{Cross Entropy} & $57.80 \pm 0.25$ & $33.85 \pm 0.13$ & $30.10 \pm 0.22$ & $19.79 \pm 0.23$ & $0.35 \pm 0.11$ & $0.00 \pm 0.00$ & $0.96 \pm 0.26$ & $0.00 \pm 0.00$ & $0.00 \pm 0.00$ \\
        \textbf{Focal} \citep{lin2017focal} & $72.70 \pm 0.19$ & $65.17 \pm 0.29$ & $62.28 \pm 0.31$ & $49.66 \pm 1.09$ & $34.14 \pm 1.05$ & $29.12 \pm 0.92$ & $38.12 \pm 1.76$ & $19.46 \pm 1.49$ & $13.44 \pm 1.73$ \\
        \textbf{LDAM} \citep{cao2019learning} & $72.50 \pm 0.15$ & $65.70 \pm 0.26$ & $63.25 \pm 0.35$ & $51.13 \pm 0.78$ & $36.86 \pm 1.03$ & $30.88 \pm 1.07$ & $40.90 \pm 1.53$ & $23.24 \pm 1.69$ & $15.69 \pm 2.13$ \\
        \textbf{Bal-Softmax} \citep{ren2020balanced} & $74.18 \pm 0.08$ & $70.48 \pm 0.55$ & $69.47 \pm 0.44$ & $56.57 \pm 0.93$ & $53.37 \pm 2.31$ & $44.24 \pm 2.83$ & $48.54 \pm 2.27$ & $45.64 \pm 3.98$ & $50.60 \pm 1.40$ \\
        \textbf{Logit-Adjust} \citep{menon2020long} & $74.08 \pm 0.05$ & $70.94 \pm 0.24$ & $69.44 \pm 0.18$ & $56.00 \pm 1.39$ & $53.93 \pm 2.46$ & $49.70 \pm 2.64$ & $47.45 \pm 2.26$ & $47.76 \pm 4.07$ & $43.26 \pm 4.69$ \\
        \textbf{Post-hoc} \citep{menon2020long} & $62.54 \pm 0.11$ & $54.84 \pm 0.15$ & $49.63 \pm 0.71$ & $35.67 \pm 0.49$ & $24.14 \pm 1.18$ & $19.00 \pm 0.68$ & $22.50 \pm 0.78$ & $7.15 \pm 1.82$ & $3.81 \pm 0.97$ \\
        \textbf{Drops} \citep{wei2023distributionally} & $73.66 \pm 0.29$ & $69.14 \pm 0.38$ & $67.15 \pm 0.17$ & $58.12 \pm 0.26$ & $47.07 \pm 0.74$ & $43.42 \pm 1.19$ & $50.85 \pm 0.49$ & $36.27 \pm 1.15$ & $32.43 \pm 1.90$ \\
        \bottomrule
    \end{tabular}
    }
    \label{tab:class_shift}

\end{table*}

Note that in the label noise learning task, the class distributes with almost balanced prior. However, in practice, the prior distribution is often long-tail distributed. Hence, the combined influence of label noise and long-tail distribution is a new and overlooked challenge presented in the literature. To facilitate the exploration of class-imbalanced learning, we tried to reduce the impact of noisy labels via selecting high-quality annotated samples as recognized by dataset curation software. Human estimation suggested a noise rate of up to 22.2\%, and 10.5\% marked as uncertain. To address this, we employed two methods to remove noisy samples: a data centric curation (Algorithm \ref{alg:docta}), which removed 26.36\% of the samples, and a learning-centric curation (Appendix Algorithm \ref{alg:learning-centric}), which removed 25\%. Combined, these methods eliminated 45.15\% of the samples, with an overlap of 6.21\% between the two approaches. We provide Clothing-ADC CLT, which could be viewed as the long-tail (class-level) distributed version of Clothing-ADC. Denote by $\rho$ the imbalanced ratio between the maximum number of samples per class and the minimum number of samples per class. In practice, we provide $\rho=10, 50, 100$ (class-level) long-tail version of Clothing-ADC. 

\vspace{0.1in}
\noindent\textbf{Benchmark Efforts:} Regarding the evaluation metric, we follow from the recently proposed metric \cite{wei2023distributionally}, which considers an objective that is based on the weighted sum of class-level performances on the test data, i.e., $\sum_{i\in[K]} g_i \text{Acc}_i$, where $\text{Acc}_i$ indicates the accuracy of the class $i$: 
\begin{align*}
   \delta\textbf{-worst  accuracy:} \qquad \min_{g\in\Delta_K}~~ \sum\nolimits_{i\in [K]}g_i
   \text{Acc}_i,
   ~~~\text{s.t. }  
   D(\mathbf{g}, \mathbf{u})\leq \delta\,.\label{eq:class_dro}
\end{align*}
Here, $\Delta_K$ denotes the $(K-1)$-dimensional probability simplex, where $K$ is the number of classes as previously defined. Let $\mathbf{u}\in\Delta_K$ be the uniform distribution, and $\mathbf{g}:=[g_1, g_2, ..., g_K]$ is the class weights. The $\delta$-worst accuracy measures the worst-case $\mathbf{g}$-weighted performance with the weights constrained to lie within the $\delta$-radius ball around the target (uniform) distribution. For any chosen divergence $D$, it reduces to the mean accuracy when $\delta=0$ and to the worst accuracy for $\delta\to\infty$. The objective interpolates between these two extremes for other values of $\delta$ and captures our goal of optimizing for variations around target priors instead of more conventional objectives of optimizing for either the average accuracy at the target prior or the worst-case accuracy.

Different from the previous dataset we used in noise learning, we use a cleaner dataset for this class-imbalance learning to avoid the distractions of noisy labels. The size of this dataset consists of 56,2263 images rather than 1M. The backbone model we use is ResNet-50. For the class distributions for different $\rho$, we include them in the Appendix. All the experiments are run for 5 times and we calculate the mean and standard deviation. With the imbalance ratio going larger, the accuracy becomes worse, which is expected for a more difficult task.

\vspace{0.1in}
\noindent\textbf{Benchmark Analysis:} Logits-Adjust and Drops are most robust methods for all $\delta$ and $\rho$ because Logit-Adjust directly target the minority class by balancing the decision boundary while Drops incorporate the DRO framework by including a term for worst-case class performance.

\section{Discussion}

\subsection{Advantages of ADC Over Traditional Data Collection}

Traditional dataset collection methods are invaluable for discovering new knowledge, particularly in fields like citizen science where expert annotation is irreplaceable. However, they require tens of thousands of human labor hours to annotate each sample \citep{van2018inaturalist, deng2009imagenet}, and even with high human effort, obtaining a clean dataset remains difficult \citep{northcutt2021pervasive}.

Our proposed ADC pipeline serves a different, complementary purpose. Rather than replacing human experts with synthetic labels from models, ADC automates the collection of existing web data and reduces label design effort from days to under an hour. By leveraging Large Language Models and image search engines, ADC enables rapid iteration during the design phase and dramatically scales the number of discoverable subclasses compared to manual brainstorming.

\subsection{How ADC Reduces Label Errors}

Despite the inevitable introduction of label noise when collecting data from the web, the ADC pipeline reduces errors in two complementary ways.

\textbf{Algorithmic Label Noise Detection.}
For applications where some label noise can be tolerated, ADC employs label noise detection methods \citep{zhu2021clusterability,zhu2023unmasking} to filter the majority of mislabeled samples. As reported in Section~\ref{sec:adc}, this reduces label noise from 22.2\% to 10.7\%, achieving a 79\% agreement rate with human curation decisions.

\textbf{Cost-Efficient Human-in-the-Loop.}
For domains requiring clean labels, ADC advocates for a targeted human review step. Unlike traditional pipelines where annotators relabel samples from scratch, ADC presents annotators with a pre-filtered set of machine-labeled candidates and asks them to select correct samples. This task is mentally easier, reduces annotator over-trust in AI answers \citep{bansal2019updates,bansal2021does}, and produces a clean set backed by both human and machine consensus \citep{liu2023humans}. We successfully applied this approach to produce a 20,000-sample clean test set for Clothing-ADC at a cost of only \$150.

\subsection{Scope of LLM Usage}

A key design principle of ADC is that LLMs are used \emph{only} during the label design phase. This deliberate restriction prevents hallucinated or inaccurate labels from propagating into the dataset, since all generated categories are reviewed by human designers before being used as search queries. Image collection and labeling is performed via commercial search engines, guaranteeing that only real, web-indexed images are included. This stands in contrast to approaches that use generative models to synthesize training images, which can introduce systematic biases and artifacts.

\section{Limitation}
While our proposed ADC pipeline demonstrates promising results for categorical labeling tasks, it has a limitation that is important to acknowledge. Currently, the pipeline is specifically designed for categorical labeling. A natural direction for future work is to expand the pipeline's scope to support a broader range of tasks, including object detection and segmentation.

\section{Conclusion}
In this paper, we introduced the Automatic Dataset Construction (ADC) pipeline, a novel approach for automating the creation of large-scale datasets with minimal human intervention. By leveraging Large Language Models for detailed class design and automated sample collection, ADC significantly reduces the time, cost, and errors associated with traditional dataset construction methods. The Clothing-ADC dataset, which comprises one million images with rich category hierarchies, demonstrates the effectiveness of ADC in producing high-quality datasets tailored for complex research tasks. Despite its advantages, ADC faces challenges such as label noise and imbalanced data distributions. We addressed these challenges with open-source tools for error detection and robust learning. Our benchmark datasets further facilitate research in these areas, ensuring that ADC remains a valuable tool for advancing machine learning model training.

\paragraph{Ethical Statement}
The Automatic Dataset Construction (ADC) pipeline emphasizes ethical data usage, relying only on publicly available sources that comply with copyright and privacy regulations.
All images collected have been returned by a ‘search engine’. All major search companies include safe search option when returning images, and we rely on their implementation.
Regarding data licensing, images are collected via the Google and Bing Image Search APIs, which index publicly available web content. This practice follows the same precedent established by widely used large-scale datasets such as LAION-5B \citep{schuhmann2022laion} and WebVision \citep{li2017webvision}, which are also assembled from publicly indexed images. We do not redistribute original image files; instead, the dataset provides image URLs and metadata. Users of the Clothing-ADC dataset are responsible for ensuring their downstream use complies with the licensing terms of individual images and applicable regulations in their jurisdiction.
Transparency in data handling and clear disclosures are prioritized, encouraging ethical considerations when using ADC, especially in sensitive domains.

\paragraph{Reproducibility Statement}

We ensure reproducibility by providing open-source code for the ADC pipeline when published. In the paper, we give detailed descriptions of data collection (Section 2), label noise mitigation (Section 3), and class balancing (Section 4). All experiments follow standard models, with full documentation of experimental setups and evaluation metrics in the Appendix C. The reproducible code is uploaded as well.

\newpage

\bibliographystyle{unsrtnat}
\bibliography{references}  

@inproceedings{wei2020combating,
  title={Combating noisy labels by agreement: A joint training method with co-regularization},
  author={Wei, Hongxin and Feng, Lei and Chen, Xiangyu and An, Bo},
  booktitle={Proceedings of the IEEE/CVF conference on computer vision and pattern recognition},
  pages={13726--13735},
  year={2020}
}

@article{chen2022improved,
  title={Improved cross entropy loss for noisy labels in vision leaf disease classification},
  author={Chen, Yipeng and Xu, Ke and Zhou, Peng and Ban, Xiaojuan and He, Di},
  journal={IET Image Processing},
  volume={16},
  number={6},
  pages={1511--1519},
  year={2022},
  publisher={Wiley Online Library}
}

@inproceedings{wei2023mitigating,
  title={Mitigating memorization of noisy labels by clipping the model prediction},
  author={Wei, Hongxin and Zhuang, Huiping and Xie, Renchunzi and Feng, Lei and Niu, Gang and An, Bo and Li, Yixuan},
  booktitle={International Conference on Machine Learning},
  pages={36868--36886},
  year={2023},
  organization={PMLR}
}

@article{natarajan2013learning,
  title={Learning with noisy labels},
  author={Natarajan, Nagarajan and Dhillon, Inderjit S and Ravikumar, Pradeep K and Tewari, Ambuj},
  journal={Advances in neural information processing systems},
  volume={26},
  year={2013}
}

@article{pruthi2020estimating,
  title={Estimating training data influence by tracing gradient descent},
  author={Pruthi, Garima and Liu, Frederick and Kale, Satyen and Sundararajan, Mukund},
  journal={Advances in Neural Information Processing Systems},
  volume={33},
  pages={19920--19930},
  year={2020}
}

@article{zhang2018generalized,
  title={Generalized cross entropy loss for training deep neural networks with noisy labels},
  author={Zhang, Zhilu and Sabuncu, Mert},
  journal={Advances in neural information processing systems},
  volume={31},
  year={2018}
}

@inproceedings{wang2019symmetric,
  title={Symmetric cross entropy for robust learning with noisy labels},
  author={Wang, Yisen and Ma, Xingjun and Chen, Zaiyi and Luo, Yuan and Yi, Jinfeng and Bailey, James},
  booktitle={Proceedings of the IEEE/CVF international conference on computer vision},
  pages={322--330},
  year={2019}
}

@article{liu2015classification,
  title={Classification with noisy labels by importance reweighting},
  author={Liu, Tongliang and Tao, Dacheng},
  journal={IEEE Transactions on pattern analysis and machine intelligence},
  volume={38},
  number={3},
  pages={447--461},
  year={2015},
  publisher={IEEE}
}

@inproceedings{ren2018learning,
  title={Learning to reweight examples for robust deep learning},
  author={Ren, Mengye and Zeng, Wenyuan and Yang, Bin and Urtasun, Raquel},
  booktitle={International conference on machine learning},
  pages={4334--4343},
  year={2018},
  organization={PMLR}
}

@inproceedings{lin2017focal,
  title={Focal loss for dense object detection},
  author={Lin, Tsung-Yi and Goyal, Priya and Girshick, Ross and He, Kaiming and Doll{\'a}r, Piotr},
  booktitle={Proceedings of the IEEE international conference on computer vision},
  pages={2980--2988},
  year={2017}
}

@inproceedings{zhu2022detecting,
  title={Detecting corrupted labels without training a model to predict},
  author={Zhu, Zhaowei and Dong, Zihao and Liu, Yang},
  booktitle={International conference on machine learning},
  pages={27412--27427},
  year={2022},
  organization={PMLR}
}

@article{menon2020long,
  title={Long-tail learning via logit adjustment},
  author={Menon, Aditya Krishna and Jayasumana, Sadeep and Rawat, Ankit Singh and Jain, Himanshu and Veit, Andreas and Kumar, Sanjiv},
  journal={arXiv preprint arXiv:2007.07314},
  year={2020}
}

@article{ren2020balanced,
  title={Balanced meta-softmax for long-tailed visual recognition},
  author={Ren, Jiawei and Yu, Cunjun and Ma, Xiao and Zhao, Haiyu and Yi, Shuai and others},
  journal={Advances in neural information processing systems},
  volume={33},
  pages={4175--4186},
  year={2020}
}

@article{cao2019learning,
  title={Learning imbalanced datasets with label-distribution-aware margin loss},
  author={Cao, Kaidi and Wei, Colin and Gaidon, Adrien and Arechiga, Nikos and Ma, Tengyu},
  journal={Advances in neural information processing systems},
  volume={32},
  year={2019}
}

@article{brodley1999identifying,
  title={Identifying mislabeled training data},
  author={Brodley, Carla E and Friedl, Mark A},
  journal={Journal of artificial intelligence research},
  volume={11},
  pages={131--167},
  year={1999}
}

@inproceedings{laurikkala2001improving,
  title={Improving identification of difficult small classes by balancing class distribution},
  author={Laurikkala, Jorma},
  booktitle={Artificial Intelligence in Medicine: 8th Conference on Artificial Intelligence in Medicine in Europe, AIME 2001 Cascais, Portugal, July 1--4, 2001, Proceedings 8},
  pages={63--66},
  year={2001},
  organization={Springer}
}

@inproceedings{xia2020robust,
  title={Robust early-learning: Hindering the memorization of noisy labels},
  author={Xia, Xiaobo and Liu, Tongliang and Han, Bo and Gong, Chen and Wang, Nannan and Ge, Zongyuan and Chang, Yi},
  booktitle={International conference on learning representations},
  year={2020}
}

@inproceedings{zhu2021clusterability,
  title={Clusterability as an alternative to anchor points when learning with noisy labels},
  author={Zhu, Zhaowei and Song, Yiwen and Liu, Yang},
  booktitle={International Conference on Machine Learning},
  pages={12912--12923},
  year={2021},
  organization={PMLR}
}

@article{zhu2023unmasking,
  title={Unmasking and improving data credibility: A study with datasets for training harmless language models},
  author={Zhu, Zhaowei and Wang, Jialu and Cheng, Hao and Liu, Yang},
  journal={arXiv preprint arXiv:2311.11202},
  year={2023}
}

@inproceedings{
wei2023distributionally,
title={Distributionally Robust Post-hoc Classifiers under Prior Shifts},
author={Jiaheng Wei and Harikrishna Narasimhan and Ehsan Amid and Wen-Sheng Chu and Yang Liu and Abhishek Kumar},
booktitle={The Eleventh International Conference on Learning Representations },
year={2023},
url={https://openreview.net/forum?id=3KUfbI9_DQE}
}

@misc{cleanlab,
  title = {Cleanlab},
  howpublished = {\url{https://cleanlab.ai/}},
}

@misc{docta,
  title = {Docta},
  howpublished = {\url{https://docta.ai/}},
}

@article{liu2020early,
  title={Early-learning regularization prevents memorization of noisy labels},
  author={Liu, Sheng and Niles-Weed, Jonathan and Razavian, Narges and Fernandez-Granda, Carlos},
  journal={Advances in neural information processing systems},
  volume={33},
  pages={20331--20342},
  year={2020}
}

@inproceedings{han2019deep,
  title={Deep self-learning from noisy labels},
  author={Han, Jiangfan and Luo, Ping and Wang, Xiaogang},
  booktitle={Proceedings of the IEEE/CVF international conference on computer vision},
  pages={5138--5147},
  year={2019}
}

@article{northcutt2021confident,
  title={Confident learning: Estimating uncertainty in dataset labels},
  author={Northcutt, Curtis and Jiang, Lu and Chuang, Isaac},
  journal={Journal of Artificial Intelligence Research},
  volume={70},
  pages={1373--1411},
  year={2021}
}

@article{lyu2023llm,
  title={Llm-rec: Personalized recommendation via prompting large language models},
  author={Lyu, Hanjia and Jiang, Song and Zeng, Hanqing and Xia, Yinglong and Luo, Jiebo},
  journal={arXiv preprint arXiv:2307.15780},
  year={2023}
}

@article{wu2023tidybot,
  title={Tidybot: Personalized robot assistance with large language models},
  author={Wu, Jimmy and Antonova, Rika and Kan, Adam and Lepert, Marion and Zeng, Andy and Song, Shuran and Bohg, Jeannette and Rusinkiewicz, Szymon and Funkhouser, Thomas},
  journal={Autonomous Robots},
  volume={47},
  number={8},
  pages={1087--1102},
  year={2023},
  publisher={Springer}
}

@inproceedings{xiao2015learning,
  title={Learning from massive noisy labeled data for image classification},
  author={Xiao, Tong and Xia, Tian and Yang, Yi and Huang, Chang and Wang, Xiaogang},
  booktitle={Proceedings of the IEEE conference on computer vision and pattern recognition},
  pages={2691--2699},
  year={2015}
}

@article{wozniak2024personalized,
  title={Personalized Large Language Models},
  author={Wo{\'z}niak, Stanis{\l}aw and Koptyra, Bart{\l}omiej and Janz, Arkadiusz and Kazienko, Przemys{\l}aw and Koco{\'n}, Jan},
  journal={arXiv preprint arXiv:2402.09269},
  year={2024}
}

@article{porsdam2023autogen,
  title={AUTOGEN: A personalized large language model for academic enhancement—Ethics and proof of principle},
  author={Porsdam Mann, Sebastian and Earp, Brian D and M{\o}ller, Nikolaj and Vynn, Suren and Savulescu, Julian},
  journal={The American Journal of Bioethics},
  volume={23},
  number={10},
  pages={28--41},
  year={2023},
  publisher={Taylor \& Francis}
}

@article{benary2023leveraging,
  title={Leveraging large language models for decision support in personalized oncology},
  author={Benary, Manuela and Wang, Xing David and Schmidt, Max and Soll, Dominik and Hilfenhaus, Georg and Nassir, Mani and Sigler, Christian and Kn{\"o}dler, Maren and Keller, Ulrich and Beule, Dieter and others},
  journal={JAMA Network Open},
  volume={6},
  number={11},
  pages={e2343689--e2343689},
  year={2023},
  publisher={American Medical Association}
}

@article{li2017webvision,
  title={Webvision database: Visual learning and understanding from web data},
  author={Li, Wen and Wang, Limin and Li, Wei and Agustsson, Eirikur and Van Gool, Luc},
  journal={arXiv preprint arXiv:1708.02862},
  year={2017}
}

@inproceedings{deng2009imagenet,
  title={Imagenet: A large-scale hierarchical image database},
  author={Deng, Jia and Dong, Wei and Socher, Richard and Li, Li-Jia and Li, Kai and Fei-Fei, Li},
  booktitle={2009 IEEE conference on computer vision and pattern recognition},
  pages={248--255},
  year={2009},
  organization={Ieee}
}

@inproceedings{
wei2022learning,
title={Learning with Noisy Labels Revisited: A Study Using Real-World Human Annotations},
author={Jiaheng Wei and Zhaowei Zhu and Hao Cheng and Tongliang Liu and Gang Niu and Yang Liu},
booktitle={International Conference on Learning Representations},
year={2022},
url={https://openreview.net/forum?id=TBWA6PLJZQm}
}

@inproceedings{patrini2017making,
  title={Making deep neural networks robust to label noise: A loss correction approach},
  author={Patrini, Giorgio and Rozza, Alessandro and Krishna Menon, Aditya and Nock, Richard and Qu, Lizhen},
  booktitle={Proceedings of the IEEE conference on computer vision and pattern recognition},
  pages={1944--1952},
  year={2017}
}

@article{li2020dividemix,
  title={Dividemix: Learning with noisy labels as semi-supervised learning},
  author={Li, Junnan and Socher, Richard and Hoi, Steven CH},
  journal={arXiv preprint arXiv:2002.07394},
  year={2020}
}

@inproceedings{wei2022smooth,
  title={To Smooth or Not? When Label Smoothing Meets Noisy Labels},
  author={Wei, Jiaheng and Liu, Hangyu and Liu, Tongliang and Niu, Gang and Sugiyama, Masashi and Liu, Yang},
  booktitle={International Conference on Machine Learning},
  pages={23589--23614},
  year={2022},
  organization={PMLR}
}

@inproceedings{jiang2018mentornet,
  title={Mentornet: Learning data-driven curriculum for very deep neural networks on corrupted labels},
  author={Jiang, Lu and Zhou, Zhengyuan and Leung, Thomas and Li, Li-Jia and Fei-Fei, Li},
  booktitle={International conference on machine learning},
  pages={2304--2313},
  year={2018},
  organization={PMLR}
}

@article{han2018co,
  title={Co-teaching: Robust training of deep neural networks with extremely noisy labels},
  author={Han, Bo and Yao, Quanming and Yu, Xingrui and Niu, Gang and Xu, Miao and Hu, Weihua and Tsang, Ivor and Sugiyama, Masashi},
  journal={Advances in neural information processing systems},
  volume={31},
  year={2018}
}

@inproceedings{lukasik2020does,
  title={Does label smoothing mitigate label noise?},
  author={Lukasik, Michal and Bhojanapalli, Srinadh and Menon, Aditya and Kumar, Sanjiv},
  booktitle={International Conference on Machine Learning},
  pages={6448--6458},
  year={2020},
  organization={PMLR}
}

@article{muller2019does,
  title={When does label smoothing help?},
  author={M{\"u}ller, Rafael and Kornblith, Simon and Hinton, Geoffrey E},
  journal={Advances in neural information processing systems},
  volume={32},
  year={2019}
}

@article{zhang2017mixup,
  title={mixup: Beyond empirical risk minimization},
  author={Zhang, Hongyi and Cisse, Moustapha and Dauphin, Yann N and Lopez-Paz, David},
  journal={arXiv preprint arXiv:1710.09412},
  year={2017}
}

@article{cheng2020learning,
  title={Learning with instance-dependent label noise: A sample sieve approach},
  author={Cheng, Hao and Zhu, Zhaowei and Li, Xingyu and Gong, Yifei and Sun, Xing and Liu, Yang},
  journal={arXiv preprint arXiv:2010.02347},
  year={2020}
}

@article{wei2020optimizing,
  title={When optimizing $ f $-divergence is robust with label noise},
  author={Wei, Jiaheng and Liu, Yang},
  journal={arXiv preprint arXiv:2011.03687},
  year={2020}
}

@inproceedings{liu2020peer,
  title={Peer loss functions: Learning from noisy labels without knowing noise rates},
  author={Liu, Yang and Guo, Hongyi},
  booktitle={International conference on machine learning},
  pages={6226--6236},
  year={2020},
  organization={PMLR}
}

@article{wei2021learning,
  title={Learning with noisy labels revisited: A study using real-world human annotations},
  author={Wei, Jiaheng and Zhu, Zhaowei and Cheng, Hao and Liu, Tongliang and Niu, Gang and Liu, Yang},
  journal={arXiv preprint arXiv:2110.12088},
  year={2021}
}

@inproceedings{song2019selfie,
title={{SELFIE}: Refurbishing Unclean Samples for Robust Deep Learning},
author={Song, Hwanjun and Kim, Minseok and Lee, Jae-Gil},
booktitle={ICML},
year={2019} }

@inproceedings{van2018inaturalist,
  title={The inaturalist species classification and detection dataset},
  author={Van Horn, Grant and Mac Aodha, Oisin and Song, Yang and Cui, Yin and Sun, Chen and Shepard, Alex and Adam, Hartwig and Perona, Pietro and Belongie, Serge},
  booktitle={Proceedings of the IEEE conference on computer vision and pattern recognition},
  pages={8769--8778},
  year={2018}
}

@inproceedings{bossard14,
  title = {Food-101 -- Mining Discriminative Components with Random Forests},
  author = {Bossard, Lukas and Guillaumin, Matthieu and Van Gool, Luc},
  booktitle = {European Conference on Computer Vision},
  year = {2014}
}

@article{liu2023humans,
  title={Do humans and machines have the same eyes? Human-machine perceptual differences on image classification},
  author={Liu, Minghao and Wei, Jiaheng and Liu, Yang and Davis, James},
  journal={arXiv preprint arXiv:2304.08733},
  year={2023}
}

@inproceedings{bansal2021does,
  title={Does the whole exceed its parts? the effect of ai explanations on complementary team performance},
  author={Bansal, Gagan and Wu, Tongshuang and Zhou, Joyce and Fok, Raymond and Nushi, Besmira and Kamar, Ece and Ribeiro, Marco Tulio and Weld, Daniel},
  booktitle={Proceedings of the 2021 CHI conference on human factors in computing systems},
  pages={1--16},
  year={2021}
}

@inproceedings{bansal2019updates,
  title={Updates in human-ai teams: Understanding and addressing the performance/compatibility tradeoff},
  author={Bansal, Gagan and Nushi, Besmira and Kamar, Ece and Weld, Daniel S and Lasecki, Walter S and Horvitz, Eric},
  booktitle={Proceedings of the AAAI Conference on Artificial Intelligence},
  volume={33},
  number={01},
  pages={2429--2437},
  year={2019}
}

@article{chawla2002smote,
  title={SMOTE: synthetic minority over-sampling technique},
  author={Chawla, Nitesh V and Bowyer, Kevin W and Hall, Lawrence O and Kegelmeyer, W Philip},
  journal={Journal of artificial intelligence research},
  volume={16},
  pages={321--357},
  year={2002}
}

@inproceedings{he2008adasyn,
  title={ADASYN: Adaptive synthetic sampling approach for imbalanced learning},
  author={He, Haibo and Bai, Yang and Garcia, Edwardo A and Li, Shutao},
  booktitle={2008 IEEE international joint conference on neural networks (IEEE world congress on computational intelligence)},
  pages={1322--1328},
  year={2008},
  organization={Ieee}
}

@inproceedings{han2005borderline,
  title={Borderline-SMOTE: a new over-sampling method in imbalanced data sets learning},
  author={Han, Hui and Wang, Wen-Yuan and Mao, Bing-Huan},
  booktitle={International conference on intelligent computing},
  pages={878--887},
  year={2005},
  organization={Springer}
}

@inproceedings{bunkhumpornpat2009safe,
  title={Safe-level-smote: Safe-level-synthetic minority over-sampling technique for handling the class imbalanced problem},
  author={Bunkhumpornpat, Chumphol and Sinapiromsaran, Krung and Lursinsap, Chidchanok},
  booktitle={Advances in Knowledge Discovery and Data Mining: 13th Pacific-Asia Conference, PAKDD 2009 Bangkok, Thailand, April 27-30, 2009 Proceedings 13},
  pages={475--482},
  year={2009},
  organization={Springer}
}

@inproceedings{mani2003knn,
  title={kNN approach to unbalanced data distributions: a case study involving information extraction},
  author={Mani, Inderjeet and Zhang, I},
  booktitle={Proceedings of workshop on learning from imbalanced datasets},
  volume={126},
  number={1},
  pages={1--7},
  year={2003},
  organization={ICML}
}

@article{tomek1976two,
  title={Two Modifications of CNN},
  author={TOMEK, I},
  journal={IEEE Trans. Systems, Man and Cybernetics},
  volume={6},
  pages={769--772},
  year={1976}
}

@inproceedings{kubat1997addressing,
  title={Addressing the curse of imbalanced training sets: one-sided selection},
  author={Kubat, Miroslav and Matwin, Stan and others},
  booktitle={Icml},
  volume={97},
  number={1},
  pages={179},
  year={1997},
  organization={Citeseer}
}

@article{jo2004class,
  title={Class imbalances versus small disjuncts},
  author={Jo, Taeho and Japkowicz, Nathalie},
  journal={ACM Sigkdd Explorations Newsletter},
  volume={6},
  number={1},
  pages={40--49},
  year={2004},
  publisher={ACM New York, NY, USA}
}

@incollection{lawrence2002neural,
  title={Neural network classification and prior class probabilities},
  author={Lawrence, Steve and Burns, Ian and Back, Andrew and Tsoi, Ah Chung and Giles, C Lee},
  booktitle={Neural networks: tricks of the trade},
  pages={299--313},
  year={2002},
  publisher={Springer}
}

@article{richard1991neural,
  title={Neural network classifiers estimate Bayesian a posteriori probabilities},
  author={Richard, Michael D and Lippmann, Richard P},
  journal={Neural computation},
  volume={3},
  number={4},
  pages={461--483},
  year={1991},
  publisher={MIT Press}
}

@inproceedings{elkan2001foundations,
  title={The foundations of cost-sensitive learning},
  author={Elkan, Charles},
  booktitle={International joint conference on artificial intelligence},
  volume={17},
  number={1},
  pages={973--978},
  year={2001},
  organization={Lawrence Erlbaum Associates Ltd}
}

@article{zhou2005training,
  title={Training cost-sensitive neural networks with methods addressing the class imbalance problem},
  author={Zhou, Zhi-Hua and Liu, Xu-Ying},
  journal={IEEE Transactions on knowledge and data engineering},
  volume={18},
  number={1},
  pages={63--77},
  year={2005},
  publisher={IEEE}
}

@inproceedings{kukar1998cost,
  title={Cost-sensitive learning with neural networks.},
  author={Kukar, Matjaz and Kononenko, Igor and others},
  booktitle={ECAI},
  volume={15},
  number={27},
  pages={88--94},
  year={1998},
  organization={Citeseer}
}

@article{tan2024democratizing,
  title={Democratizing Large Language Models via Personalized Parameter-Efficient Fine-tuning},
  author={Tan, Zhaoxuan and Zeng, Qingkai and Tian, Yijun and Liu, Zheyuan and Yin, Bing and Jiang, Meng},
  journal={arXiv preprint arXiv:2402.04401},
  year={2024}
}

@inproceedings{chang2017revolt,
  title={Revolt: Collaborative crowdsourcing for labeling machine learning datasets},
  author={Chang, Joseph Chee and Amershi, Saleema and Kamar, Ece},
  booktitle={Proceedings of the 2017 CHI conference on human factors in computing systems},
  pages={2334--2346},
  year={2017}
}

@inproceedings{kulesza2014structured,
  title={Structured labeling for facilitating concept evolution in machine learning},
  author={Kulesza, Todd and Amershi, Saleema and Caruana, Rich and Fisher, Danyel and Charles, Denis},
  booktitle={Proceedings of the SIGCHI Conference on Human Factors in Computing Systems},
  pages={3075--3084},
  year={2014}
}

@inproceedings{ramaswamy2023overlooked,
  title={Overlooked factors in concept-based explanations: Dataset choice, concept learnability, and human capability},
  author={Ramaswamy, Vikram V and Kim, Sunnie SY and Fong, Ruth and Russakovsky, Olga},
  booktitle={Proceedings of the IEEE/CVF Conference on Computer Vision and Pattern Recognition},
  pages={10932--10941},
  year={2023}
}

@inproceedings{bansal2021most,
  title={Is the most accurate ai the best teammate? optimizing ai for teamwork},
  author={Bansal, Gagan and Nushi, Besmira and Kamar, Ece and Horvitz, Eric and Weld, Daniel S},
  booktitle={Proceedings of the AAAI Conference on Artificial Intelligence},
  volume={35},
  number={13},
  pages={11405--11414},
  year={2021}
}

@inproceedings{han2021iterative,
  title={Iterative human-in-the-loop discovery of unknown unknowns in image datasets},
  author={Han, Lei and Dong, Xiao and Demartini, Gianluca},
  booktitle={Proceedings of the AAAI Conference on Human Computation and Crowdsourcing},
  volume={9},
  pages={72--83},
  year={2021}
}

@inproceedings{sheng2008get,
  title={Get another label? improving data quality and data mining using multiple, noisy labelers},
  author={Sheng, Victor S and Provost, Foster and Ipeirotis, Panagiotis G},
  booktitle={Proceedings of the 14th ACM SIGKDD international conference on Knowledge discovery and data mining},
  pages={614--622},
  year={2008}
}

@article{krizhevsky2009learning,
  title={Learning multiple layers of features from tiny images},
  author={Krizhevsky, Alex and Hinton, Geoffrey and others},
  year={2009},
  publisher={Toronto, ON, Canada}
}

@inproceedings{liu2015faceattributes,
  title = {Deep Learning Face Attributes in the Wild},
  author = {Liu, Ziwei and Luo, Ping and Wang, Xiaogang and Tang, Xiaoou},
  booktitle = {Proceedings of International Conference on Computer Vision (ICCV)},
  month = {December},
  year = {2015} 
}

@inproceedings{he2016deep,
  title={Deep residual learning for image recognition},
  author={He, Kaiming and Zhang, Xiangyu and Ren, Shaoqing and Sun, Jian},
  booktitle={Proceedings of the IEEE conference on computer vision and pattern recognition},
  pages={770--778},
  year={2016}
}

@article{papernot2018deep,
  title={Deep k-nearest neighbors: Towards confident, interpretable and robust deep learning},
  author={Papernot, Nicolas and McDaniel, Patrick},
  journal={arXiv preprint arXiv:1803.04765},
  year={2018}
}

@article{schuhmann2022laion,
  title={Laion-5b: An open large-scale dataset for training next generation image-text models},
  author={Schuhmann, Christoph and Beaumont, Romain and Vencu, Richard and Gordon, Cade and Wightman, Ross and Cherti, Mehdi and Coombes, Theo and Katta, Aarush and Mullis, Clayton and Wortsman, Mitchell and others},
  journal={Advances in Neural Information Processing Systems},
  volume={35},
  pages={25278--25294},
  year={2022}
}

@article{northcutt2021pervasive,
  title={Pervasive label errors in test sets destabilize machine learning benchmarks},
  author={Northcutt, Curtis G and Athalye, Anish and Mueller, Jonas},
  journal={arXiv preprint arXiv:2103.14749},
  year={2021}
}

@article{huang2023survey,
  title={A survey on hallucination in large language models: Principles, taxonomy, challenges, and open questions},
  author={Huang, Lei and Yu, Weijiang and Ma, Weitao and Zhong, Weihong and Feng, Zhangyin and Wang, Haotian and Chen, Qianglong and Peng, Weihua and Feng, Xiaocheng and Qin, Bing and others},
  journal={arXiv preprint arXiv:2311.05232},
  year={2023}
}

@article{xu2024hallucination,
  title={Hallucination is inevitable: An innate limitation of large language models},
  author={Xu, Ziwei and Jain, Sanjay and Kankanhalli, Mohan},
  journal={arXiv preprint arXiv:2401.11817},
  year={2024}
}

\newpage

\appendix

\section*{Appendix}
The appendix is organized as follows:

\squishlist
\item Appendix \ref{app:details_adc} includes additional detailed algorithms in the Automatic-Dataset-Construction pipeline.
\item Appendix \ref{app:statistics} contains dataset statistics and more exploratory data analysis of Clothing ADC.
\item Appendix \ref{app:exp_details} includes experiment details of our benchmark on label noise detection, label noise learning, and class-imbalanced learning.
\squishend

\subsection*{Broader impacts}

Our paper introduces significant advancements in dataset construction methodologies, particularly through the development of the Automatic Dataset Construction (ADC) pipeline:

\squishlist
\item \textbf{Reduction in Human Workload:} ADC automates the process of dataset creation, significantly reducing the need for manual annotation and thereby decreasing both the time and costs associated with data curation.
\item \textbf{Enhanced Data Quality for Research Communities:} ADC provides high-quality, tailored datasets with minimal human intervention. This provides researchers with datasets in the fields of label noise detection, label noise learning, and class-imbalanced learning, for exploration as well as fair comparisons.
\item \textbf{Support for Customized LLM Training:} The ability to rapidly generate and refine datasets tailored for specific tasks enhances the training of customized Large Language Models (LLMs), increasing their effectiveness and applicability in specialized applications.
\squishend
Furthermore, the complementary software developed alongside ADC enhances these impacts:
\squishlist
\item \textbf{Data Curation and Quality Control:} The software aids in curating and cleaning the collected data, ensuring that the datasets are of high quality that could compromise model training.
\item \textbf{Robust Learning Capabilities:} It incorporates methods for robust learning with collected data, addressing challenges such as label noise and class imbalances. This enhances the reliability and accuracy of models trained on ADC-constructed datasets.
\squishend
Together, ADC and its accompanying software significantly advance the capabilities of machine learning researchers and developers by providing efficient tools for high-quality customized data collection, and robust training.

\subsection*{Limitations}
\label{sec:limitation}
 While ensuring the legal and ethical use of datasets, including compliance with copyright laws and privacy concerns, is critical, our initial focus is on legally regulated and license-friendly data sources available through platforms like Google or Bing. Addressing these ethical considerations is beyond the current scope but remains an essential aspect of dataset usage. 
 
 Besides, similar to Traditional-Dataset-Construction (TDC), Automatic-Dataset-Construction (ADC) is also unable to guarantee fully accurate annotations.
\newpage
\section{Detailed algorithms in the generation of Automatic-Dataset-Construction}\label{app:details_adc}

\subsection{Image data collection in ADC}
\label{alg:collect_data}

Algorithm~\ref{alg:image-data-collection} details our automated image collection process. The algorithm takes dataset design attributes as input and systematically generates search queries by combining colors, patterns, and materials with target categories. It then launches distributed searches across multiple image search APIs to collect relevant samples at scale.

\begin{algorithm}[htbp]
    \caption{Image Data Collection in ADC}
    \label{alg:image-data-collection}
    \begin{algorithmic}[1]
    \Procedure{ImageDataCollection}{}
        \State \textbf{Part A: }Get attributes from dataset design
        \State $\mathbf{attributes} \leftarrow$ Step 1 Dataset Design
        \State $\mathbf{categories} \leftarrow [\text{"sweater"}, \text{"shirt"}, \text{"pants"}, ...]$ 
        \State \Comment{List of categories}
        \State $\mathbf{target\_category} \leftarrow \text{"sweater"}$ 
        \State \Comment{Target category (e.g. "sweater")}
        \State $\mathbf{attributes} \leftarrow attributes[\mathbf{target\_category}]$ \State \Comment{Get attributes for target category}
        \State $colors, patterns, materials \leftarrow \mathbf{attributes}[\text{"color"}],$ 
        \State \hspace{13em} $\mathbf{attributes}[\text{"pattern"}],$
        \State \hspace{13em} $\mathbf{attributes}[\text{"material"}]$
    
        \State \textbf{Part B: }Create search queries
    
        \State $\mathbf{search\_queries} \leftarrow \{ \, c + p + m + \mathbf{target\_category} \mid$
        \State \hspace{9em} $c \in \mathbf{colors},$
        \State \hspace{9em} $p \in \mathbf{patterns},$
        \State \hspace{9em} $m \in \mathbf{materials}\}$ 
        \State \Comment{(e.g. "beige fisherman cotton sweater")}
    
        \State \textbf{Part C: }Launch distributed image search
        \State $image\_data \leftarrow$ distributed\_search($search\_queries,$
        \State \hspace{6em} $api=\text{Google\_Images} \mid \text{Bing\_Images},$
        \State \hspace{6em} $n\_process=30)$
    
    \EndProcedure
    \end{algorithmic}
    \end{algorithm}

\subsection{Learning-centric curation method in ADC}

Algorithm~\ref{alg:learning-centric} presents our learning-centric curation approach that leverages the early-learning memorization behavior of neural networks. This method trains a classifier for only 1-2 epochs and uses the model's confidence scores to identify and filter out potentially mislabeled samples, removing the bottom 25\% of samples with the lowest confidence scores.

\begin{algorithm}[htbp]
    \caption{Learning-centric curation (early-learning memorization behavior)}
    \label{alg:learning-centric}
    \begin{algorithmic}[1]
    \Procedure{EarlyStopCE}{noisyDataset, percentage=$25\%$}
        \State \textbf{Part A: }Train classifier over the dataset and apply early stopping
        \State $\mathcal{D} \leftarrow$ Load training data \Comment{(images and labels)}
        \State $model \leftarrow$ Initialize neural network model \Comment{(e.g. ResNet)}
        \State $loss\_fn \leftarrow$ Define loss function \Comment{(e.g. cross-entropy)}
        \State $optimizer \leftarrow$ Choose optimizer \Comment{(e.g. SGD, Adam)}
        \For{$epoch = 1$ to $E \in \{1, 2\}$}
            \State $model \leftarrow$ Trainer($\mathcal{D}, loss\_fn, optimizer$) 
        \EndFor
    
        \State \textbf{Part B: }Record predictions and confidence levels
        \For{$batch$ in $\mathcal{D}$}
            \State $images \leftarrow$ Get batch of images
            \State $outputs \leftarrow$ Forward pass: $model(images)$
            \State $confidence \leftarrow$ Get confidence levels: $\text{softmax}(outputs)$
            
        \EndFor
    
        \State \textbf{Part C: }Remove samples with lowest $x\%$ confidence level
        \State $threshold$ 
        \State \hspace{1em}$\leftarrow$ Calculate threshold: $\text{percentile}(confidence, 100 - x)$
        \State $\mathcal{D} \leftarrow$ Filter out samples with confidence below $threshold$
    
        \State \textbf{Return} $\mathcal{D}$
    
    \EndProcedure
    \end{algorithmic}
    \end{algorithm}

 \section{Dataset statistics in Clothing-ADC}\label{app:statistics}

\subsection{Collected Clothing ADC dataset}
Our collected Clothing-ADC dataset can be found here: \href{https://huggingface.co/datasets/mikelmh025/ClothingADC}{\textbf{[Link to HuggingFace]}}.

\subsection{Attributes candidates in Clothing-ADC}
Our automated dataset creation pipeline is capable of generating numerous designs per attribute, as shown in Table \ref{Tab:dataset_attributes_full}. This table provides a detailed list of designs generated by our pipeline, from which we selected a subset to include in our dataset.

\begin{table*}[t]
\resizebox{\textwidth}{!}{%
\begin{tabular}{|lll|lll|llllll|}
\hline
\multicolumn{3}{|c|}{Color}                        & \multicolumn{3}{c|}{Material}                   & \multicolumn{6}{c|}{Pattern}                                                                         \\ \hline
Animal print  & Gold           & Pastel            & Acrylic     & Lace            & Tulle           & Abstract        & Camouflage    & Fishnet       & Leather         & Printed       & Thongs           \\
Beige         & Gray           & Peach             & Alpaca      & Leather         & Tweed           & Abstract Floral & Chalk stripe  & Floral        & Logo            & Quilted       & Tie-Dye          \\
Black         & Green          & Pink              & Angora      & Lightweight     & Twill           & Animal Print    & Check         & Floral print  & Low rise        & Reversible    & Tie-dye          \\
Blue          & Grey           & Plum              & Bamboo      & Linen           & Velvet          & Animal print    & Checkered     & Fringe        & Mesh            & Ribbed        & Toile            \\
Blush Pink    & Heather        & Purple            & Breathable  & Mesh            & Viscose         & Aran            & Chevron       & G-strings     & Military        & Ripples       & Trench           \\
Bright Red    & Ivory          & Red               & Cashmere    & Microfiber      & Water-resistant & Argyle          & Color block   & Galaxy        & Mock turtleneck & Satin         & Tribal           \\
Brown         & Khaki          & Rich Burgundy     & Chambray    & Modal           & Windproof       & Aztec           & Colorblock    & Garter Stitch & Mosaic          & Scales        & Tuck stitch      \\
Burgundy      & Lavender       & Royal Blue        & Chiffon     & Mohair          & Wool            & Basket check    & Cotton        & Garter stitch & Moss stitch     & Seamless      & Tweed            \\
Burnt Orange  & Light Grey     & Rust              & Corduroy    & Neoprene        & acrylic         & Basket rib      & Cropped       & Geometric     & Moto            & Seed stitch   & Twill            \\
Champagne     & Maroon         & Rustic Orange     & Cotton      & Nylon           & bamboo          & Basket weave    & Damask        & Gingham       & Nailhead        & Shadow stripe & Vintage-inspired \\
Charcoal      & Metallic       & Sage              & Crochet     & Organza         & cotton          & Basketweave     & Denim         & Glen check    & Nehru           & Sharkskin     & Waterproof       \\
Charcoal Grey & Mustard        & Silver            & Denim       & PVC             & hemp            & Batik           & Diagonal grid & Gradient      & Nordic          & Sherpa        & Windowpane       \\
Cream         & Mustard Yellow & Soft Pink         & Down        & Polyester       & linen           & Bikini          & Diamond       & Graphic       & Ombre           & Silk          &                  \\
Cream White   & Navy           & Striped           & Embroidered & Rayon           & lycra           & Birdseye        & Ditsy         & Grid          & Oversized       & Slip Stitch   &                  \\
Dark Plum     & Navy Blue      & Tan               & Flannel     & Reflective      & modal           & Blazer          & Dogtooth      & Herringbone   & Oxford          & Slip stitch   &                  \\
Deep Blue     & Neon           & Teal              & Fleece      & Ripstop         & nylon           & Bomber          & Embossed      & High waisted  & Paisley         & Solid         &                  \\
Deep Purple   & Nude           & Turquoise         & Fringe      & Satin           & polyester       & Boxer briefs    & Embroidered   & Honeycomb     & Peacoat         & Striped       &                  \\
Earthy Beige  & Olive          & Vibrant Turquoise & Fur         & Silk            & rayon           & Briefs          & Emoji         & Houndstooth   & Pin Dot         & Stripes       &                  \\
Floral        & Olive Green    & Warm Brown        & Gore Tex    & Softshell       & silk            & Brioche         & Entrelac      & Ikat          & Pinstripe       & Studded       &                  \\
Forest Green  & Orange         & White             & Gore-Tex    & Spandex         & spandex         & Broken rib      & Eyelet        & Intarsia      & Plaid           & Suede         &                  \\
Fuchsia       & Pale Yellow    & Yellow            & Hemp        & Suede           & tencel          & Broken stripe   & Fair Isle     & Jacquard      & Polka Dot       & Tartan        &                  \\
              &                & lilac             & Insulated   & Synthetic       & viscose         & Cable           & Fibonacci     & Knit and Purl & Polka dot       & Teddy         &                  \\
              &                &                   & Jersey      & Synthetic Blend & wool            & Cable knit      & Fisherman     & Lace          & Prince of Wales & Textured      &                  \\
              &                &                   & Knit        & Tencel          &                 &                 &               &               &                 &               &                 \\ \hline
\end{tabular}}
\caption{The union of attributes across all clothing types in Clothing-ADC dataset.}
\label{Tab:dataset_attributes_full}
\end{table*}

\subsection{Human-in-the-Loop curation for ClothingADC testset}
\label{sec:appendix test set collection}

\begin{figure*}[t]
  \centering
  \includegraphics[width=\linewidth]{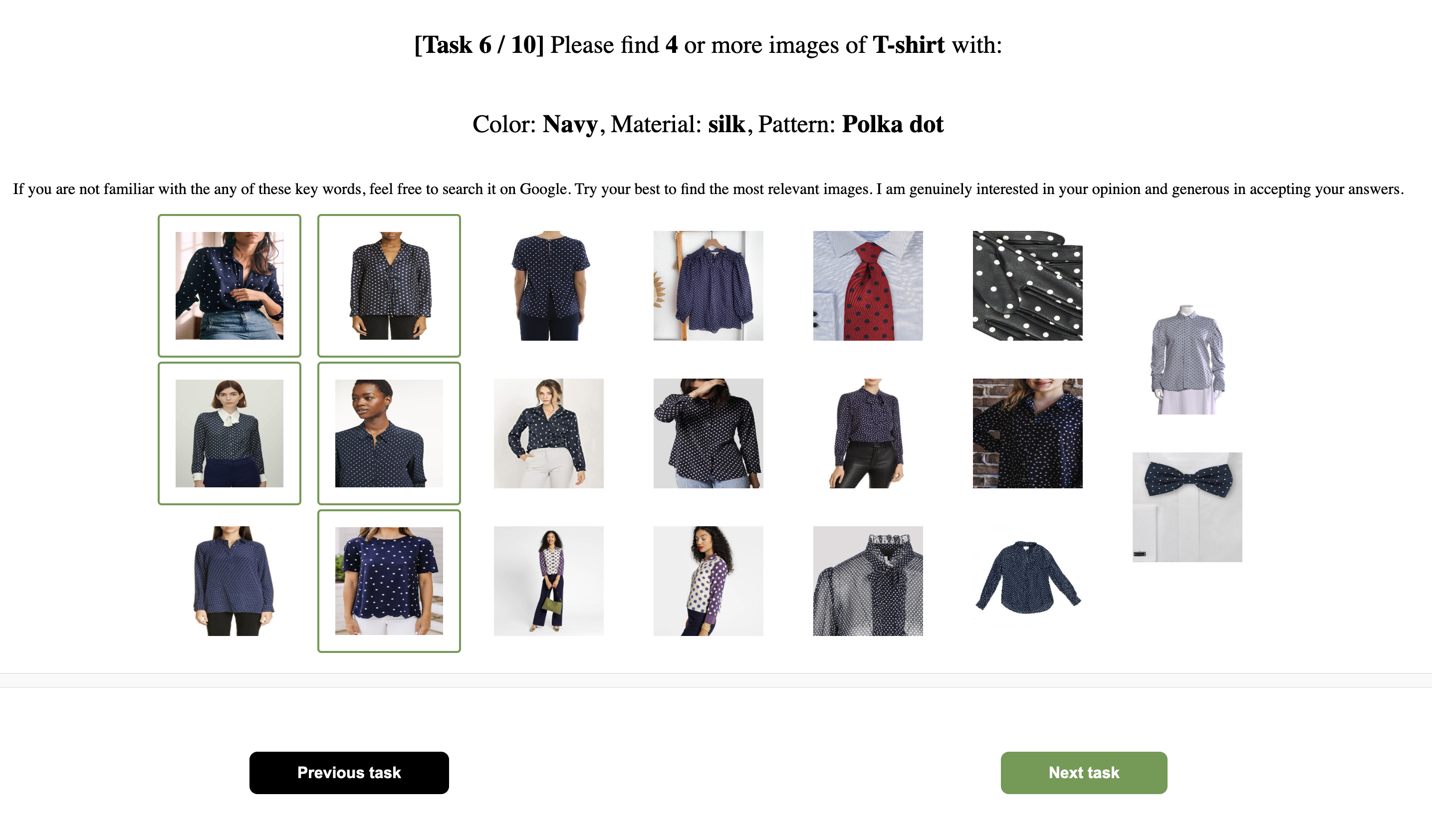}
  \caption{Collection of Clothing-ADC test set: A filtering task to the worker instead of annotation from scratch.}
  \label{fig:adc_testset_worker_page}
  \vspace{0.85in}
\end{figure*}

\begin{figure}[H]
  \centering
  \begin{subfigure}[t]{0.49\linewidth}
    \centering
    \includegraphics[width=\linewidth]{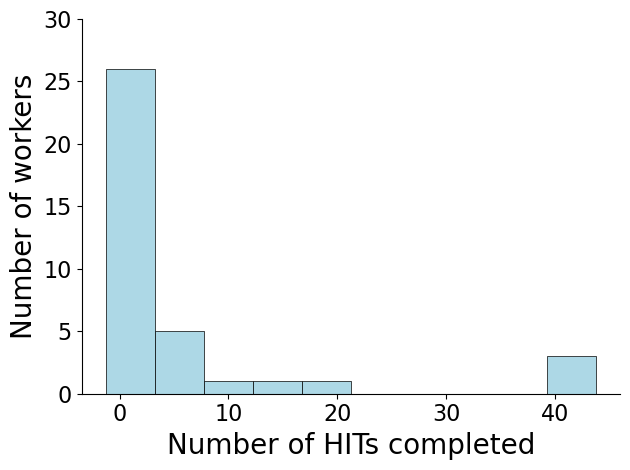}
    \caption{Distribution of the HITs completed per worker}
  \end{subfigure}
  \begin{subfigure}[t]{0.49\linewidth}
    \centering
    \includegraphics[width=\linewidth]{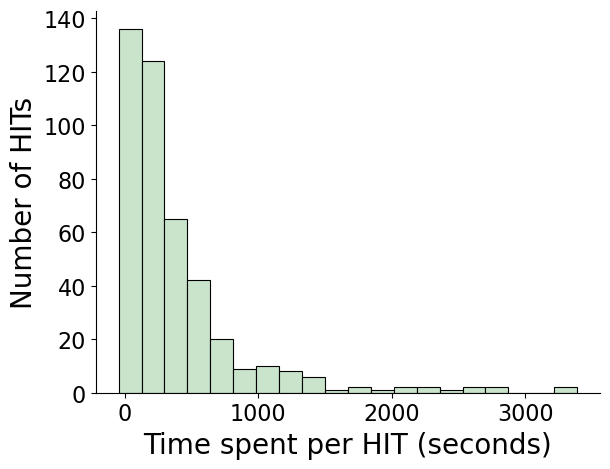}
    \caption{Distribution of work time in seconds per HIT.}
  \end{subfigure}
  \caption{The behaviors of workers in the creation of test set.}
  \label{fig:testset_creation_behaviors}
\end{figure}

Our automated dataset collection pipeline enabled us to create a large, noisy labeled dataset. We asked annotators to select the best-fitting options from a range of samples, as shown in Figure \ref{fig:adc_testset_worker_page}, with each task including at least 4 samples and workers completing 10 tasks per HIT at a cost of \$0.15 per task, totaling \$150 estimated wage of \$2.5-3 per hour, and after further cleaning the label noise, we ended up with 20,000 samples in our test set. To participate, workers had to meet specific requirements, including being Master workers, having a HIT Approval Rate above 85\%, and having more than 500 approved HITs, with the distribution of worker behavior shown in Figure \ref{fig:testset_creation_behaviors}.

\subsection{Cost analysis for ClothingADC Human-in-the-Loop data curation}
\label{sec:appendix human curation cost analysis}
When clean data is required, we recommend combining human involvement with algorithmic approaches to ensure high accuracy. We collected 20,000 samples for both the test set and evaluation set, ensuring a robust and reliable dataset. 

We evaluate human effort in Table \ref{tab:cost_testset}. We used the number of mouse clicks required for each label, excluding overhead costs due to different layout designs across datasets. While other metrics like time spent or monetary cost could be used within the same dataset, they are not easily comparable across datasets with different setups and participants.

\begin{table*}[!htb]
    \centering
    \scalebox{0.8}{
    \begin{tabular}{|l|c|c|c|c|c|c|}
        \hline
        \multirow{2}{*}{Dataset} & \multirow{2}{*}{Class Count} & \multirow{2}{*}{Noise Rate} & Label & Cost per & \multirow{2}{*}{Total Cost (\$)} & \multirow{2}{*}{Samples Collected} \\
        & & & per Sample & Label (Click) & & \\
        \hline
        ClothingADC Testset & 12k & Clean & 4 & \textbf{0.25} & \$150 / 150 & 20k / 20k \\
        \hline
        Cifar-10 N & 10 & $\sim$18\% & 1 & 3 & \$450 & 50k \\
        \hline
        Cifar-100 N & 100 & $\sim$40\% & 1 & 1 & \$700 & 50k \\
        \hline
        Cifar-10 H & 10 & 5\% & 1 & 50 & \$3,856.5 & 20k \\
        \hline
    \end{tabular}
    }
    \caption{Human Effort Comparison with Existing Label Noise Datasets.}
    \label{tab:cost_testset}
\end{table*}

\subsection{"Clean set" from Traditional methods is not always clean}
The noise rate in the manually annotated dataset iNaturalist is close to 0, suggesting that traditional methods requiring experts are more robust than our proposed ADC pipeline. However, we would like to cite \cite{northcutt2021pervasive} that even well-curated and widely-adopted “clean” test datasets, which have invested significant effort in ensuring data quality, may still contain errors \footnote{https://labelerrors.com/}. This highlights that achieving a 0\% noise rate is extremely challenging, even with expert annotation. The table below is the evidence of such observations (from Table 2 in \cite{northcutt2021pervasive}).

\begin{table}[]
    \centering
    \begin{tabular}{|l|l|l|}
        \hline
        Dataset (Test Set) & Size & \% Error \\
        \hline
        MNIST & 10000 & 0.15 \\
        CIFAR-10 & 10000 & 0.54 \\
        CIFAR-100 & 10000 & 5.85 \\
        Caltech-256 & 29780 & 1.84 \\
        ImageNet & 50000 & 5.83 \\
        QuickDraw & 50426266 & 10.12 \\
        20News & 7532 & 1.09 \\
        IMDB & 25000 & 2.90 \\
        Amazon Reviews & 9996437 & 3.90 \\
        AudioSet & 20371 & 1.35 \\
        \hline
    \end{tabular}
    \caption{Error comparison across datasets (from Table 2 in \cite{northcutt2021pervasive})}
    \label{tab:my_label}
\end{table}

Moreover, a “fully-cleaned” set typically consumes much more time and money. When the budget is limited, the annotation accuracy is much lower. For example, the collection of CIFAR-10N \cite{wei2022learning}, where each training image of CIFAR-10 (a relatively easy 10-class classification) is assigned to 3 independent annotators. To collect 3 annotations for each of the 50K images, it takes >2 days and >1000 dollars on Amazon Mturk. However, the overall annotation error is approximately 18\%. As for CIFAR-100N \cite{wei2022learning}, this is a much more challenging task where each annotator is requested to find out the most relevant label for each image among 100 classes (50K images in all). It takes >2 days and > 800 dollars on Amazon Mturk. However, the overall annotation error is approximately 40\%.

\section{Experiment details}\label{app:exp_details}





\subsection{Distribution of Human Votes for Label Noise Evaluation}
\label{sec:appendix label noise eval}

\begin{figure*}[!htb]
  \centering
  \includegraphics[width=\linewidth]{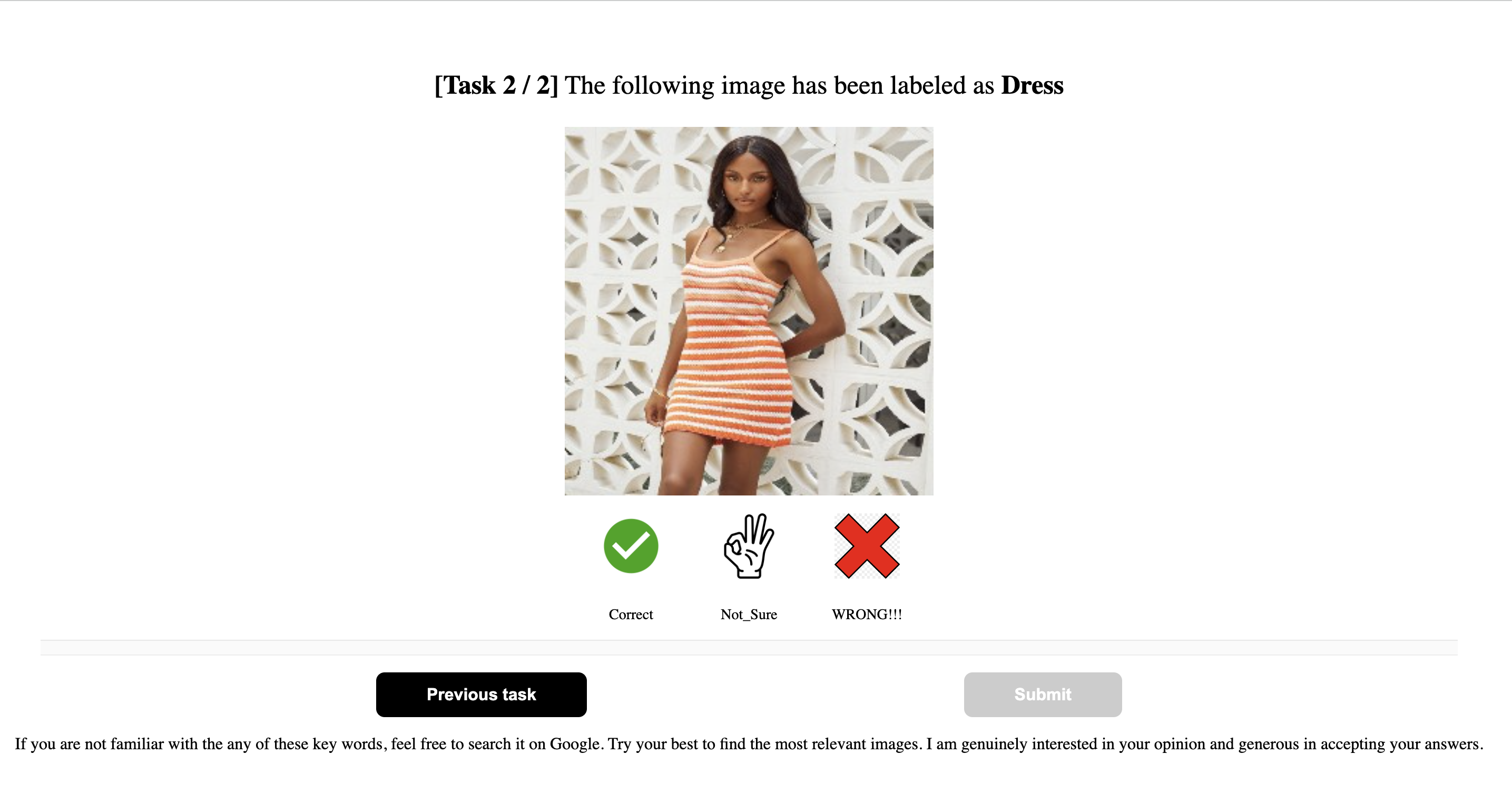}
  \caption{Label noise evaluation worker page}
  \label{fig:label_noise_worker_page}
\end{figure*}

On the annotation page, we presented the image and its original label to the worker and asked if they believed the label was correct (Figure \ref{fig:label_noise_worker_page}). They input their evaluation by clicking one of three buttons. Note that we encouraged workers to categorize acceptable samples as "unsure". The resulting distribution is shown in Table \ref{tab:label_noise_human_votes}. Using a simple majority vote aggregation, we found that the noise rate in our dataset is 22.15\%. However, if a higher level of certainty is required for clean labels, we can apply a more stringent aggregation method, considering more samples as mislabeled. In the extreme case where any doubts from any of the three annotators can disqualify a sample, our automatically collected dataset still retains 61.25\% of its samples. 

For the label noise evaluation task, we utilized a subset of 20,000 samples from the Clothing-ADC dataset, collecting three votes from unique workers for each sample. Each Human Intelligence Task (HIT) included 20 samples and cost \$0.05. To participate, workers had to meet the following requirements: (1) be Master workers, (2) have a HIT Approval Rate above 85\%, and (3) have more than 500 approved HITs. The total cost for this task was \$150, estimated wage of \$2.5-3 per hour.

\begin{figure}[!htb]
  \centering
  \begin{subfigure}[t]{0.47\linewidth}
    \centering
    \includegraphics[width=\linewidth]{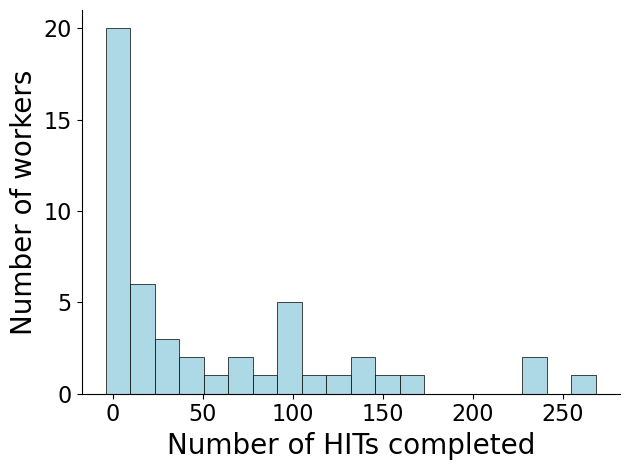}
    \caption{Distribution of the HITs completed per worker}
  \end{subfigure}
  \begin{subfigure}[t]{0.47\linewidth}
    \centering
    \includegraphics[width=\linewidth]{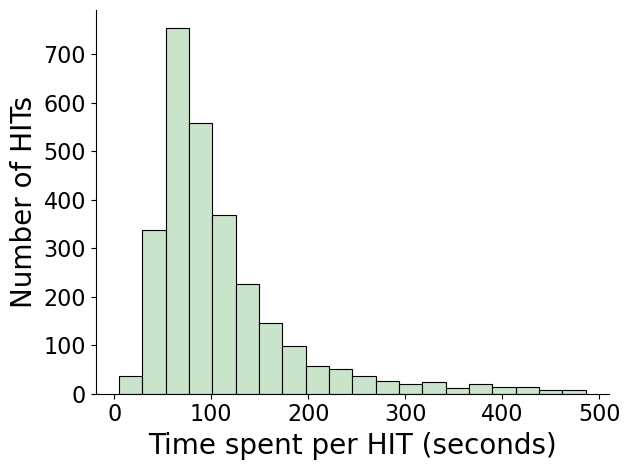}
    \caption{Distribution of work time in seconds per HIT.}
  \end{subfigure}
  \caption{The behaviors of workers in the collection of label noise evaluation.}
  \label{fig:noise_eval_behaviors}
\end{figure}

We show the distribution of worker behavior during the noise evaluation task in Figure \ref{fig:noise_eval_behaviors}. Figure \ref{fig:noise_eval_behaviors}(a) shows the distribution of the amount of HIT completed per worker while neglecting ids with 1-2 submissions. There is a total of 49 unique workers. Figure \ref{fig:noise_eval_behaviors}(b) shows the distribution of time spent per HIT.

\begin{table}[!htb]
    \caption{\textbf{Distribution of Human Votes for Label Noise Evaluation}: We employed human annotators to evaluate a subset of 20,000 samples from our collected dataset, with each sample receiving three votes from distinct annotators. }
    \centering
    \begin{tabular}{lc}
        \toprule
        \rowcolor{blue!10}\textbf{Human Votes} & \textbf{Percentage} \\
        \midrule
        Yes, Yes, Yes & 61.25\% \\
        Yes, Yes, Unsure & 6.10\% \\
        Yes, Yes, No & 10.50\% \\
        \midrule
        Else & 22.15\% \\
        \bottomrule
    \end{tabular}
    \label{tab:label_noise_human_votes}
\end{table}

\newpage
\subsection{Noisy learning and class imbalance learning benchmark implementation details}

Our code refers to zip file in supplementary material.


\begin{lstlisting}[language=Python, caption={How to load data. Line 1 loads the full set of our dataset. Line 2 and Line 3 load the tiny version of our dataset. Line 4 creates the validation set. Line 5 creates the testing set. Line 11 to Line 20 create the data loader.}]
train_set = Clothing1mPP(root, image_size, split="train")
tiny_set_ids = train_set.get_tiny_ids(seed=0)
tiny_train_set = Subset(train_set, tiny_set_ids) # Get the tiny version of the dataset
val_set = Clothing1mPP(
    root, image_size, split="val", pre_load=train_set.data_package
)
test_set = Clothing1mPP(
    root, image_size, split="test", pre_load=train_set.data_package
)

train_loader = DataLoader(
    train_set, batch_size=batch_size, shuffle=True, num_workers=num_workers
)
tiny_train_loader = DataLoader(
    tiny_train_set, batch_size=batch_size, shuffle=True, num_workers=num_workers
)
val_loader = DataLoader(
    val_set, batch_size=batch_size, shuffle=False, num_workers=num_workers
)
test_loader = DataLoader(
    test_set, batch_size=batch_size, shuffle=False, num_workers=num_workers
)
\end{lstlisting}


\begin{lstlisting}[language=bash,caption={The example of the command we use to run the algorithm in one line}]
python examples/main.py --config configs/Clothing1MPP/default.yaml # Run Cross Entropy
python examples/main_peer.py --config configs/Clothing1MPP/default.yaml # Run Peer Loss
python examples/main_jocor.py --config configs/Clothing1MPP/default_jocor.yaml # Run Jocor
python examples/main_coteaching.py --config configs/Clothing1MPP/default_coteaching.yaml # Run Co-teaching
python examples/main_drops.py --config configs/Clothing1MPP/default_drops.yaml # Run drops
\end{lstlisting}

\begin{lstlisting}[language=yaml, caption=The example of YAML config file]
inherit_from: configs/default.yaml
data: &data_default
  root: '/root/cloth1m_data_v3' 
  image_size: 256
  dataset_name: "clothing1mpp"
  imbalance_factor: 1 # 1 means no imbalance
  tiny: False

train: &train
  num_workers: 8
  loss_type: 'ce'
  loop_type: 'default' # 'default','peer','drops'
  epochs: 20
  global_iteration: 999999999
  batch_size: 64
  # scheduler_T_max: 40
  scheduler_type: 'step'
  scheduler_gamma: 0.8
  scheduler_step_size: 2
  print_every: 100
  learning_rate: 0.01

general:
  save_root: './results/'
  whip_existing_files: True # Whip exisitng files
  logger:
    project_name: 'Clothing1MPP'
    frequency: 200

model: &model_default
  name: "resnet50"
  pretrained_model: 'IMAGENET1K_V1'
  cifar: False

test: &test_defaults
  <<: *train
\end{lstlisting}

\begin{table*}[!htb]
    \centering
    \resizebox{\textwidth}{!}{%
    \begin{tabular}{c|cc}
    \toprule
        imbalance ratio ($\rho$) & Class Distribution & Total Number \\
    \midrule
        10 & [39297, 31875, 25854, 20971, 17010, 13797, 11191, 9078, 7363, 5972, 4844, 3929] & 191181 \\
        20 & [39297, 27536, 19295, 13520, 9474, 6638, 4652, 3259, 2284, 1600, 1121, 785] & 129461 \\
        100 & [39297, 25854, 17010, 11191, 7363, 4844, 3187, 2097, 1379, 907, 597, 392] & 114118 \\
    \bottomrule
    \end{tabular}}
    \caption{The class distribution for different imbalance ratio}
    \vspace{-0.3in}
    \label{tab:class_imbalance}
\end{table*}

\subsection{Label noise detection benchmark}
\label{sec:noise_detection}
We run four baselines for label noise detection, including CORES \cite{cheng2020learning}, confident learning \cite{northcutt2021confident}, deep $k$-NN \cite{papernot2018deep} and Simi-Feat \cite{zhu2022detecting}. All the experiment is run for one time following \cite{cheng2020learning, zhu2022detecting}.

The experiment platform we run is a 128-core AMD EPYC 7742 Processor CPU and the memory is 128GB. The GPU we use is a single NVIDIA A100 (80GB) GPU. For the dataset, we used human annotators to evaluate whether the sample has clean or noisy label as mentioned in Appendix \ref{sec:appendix label noise eval}. We aggressively eliminates human uncertainty factors and only consider the case with unanimous agreement as a clean sample, and everything else as noisy samples. The backbone model we use is ResNet-50 \cite{he2016deep}. 
For all the baselines, the parameters we use are the same as the original paper except the data loader. We skip the label corruption and use the default value from the original repository. 
For CORES, the cores loss whose value is smaller than 0 is regarded as the noisy sample. 
For confidence learning, we use the repository\footnote{https://github.com/cleanlab/cleanlab} from the clean lab and the default hyper-parameter.
For deep $k$-NN, the $k$ we set is 100.
For SimiFeat, we set $k$ as 10 and the feature extractor is CLIP.

\subsection{Label noise learning benchmark}
\label{sec:noise_learning}
The platform we use is the same as label noise detection. The backbone model we use is ResNet-50 \cite{he2016deep}. For the full dataset, we run the experiment for 1 time. For the tiny dataset, we run the experiments for 3 times. The tiny dataset is sampled from the full set whose size is 50. The base learning rate we use is 0.01. The base number of epochs is 20. The hyper-parameters for each baseline method are as follows.
For \textbf{backward and forward correction}, we train the model using cross-entropy (CE) loss for the first 10 epochs. We estimate the transition matrix every epoch from the 10th to the 20th epoch. For the \textbf{positive and negative label smoothing}, the smoothed labels are used at the 10th epoch. The smooth rates of the positive and negative are 0.6 and -0.2. Similarly, for \textbf{peer loss}, we train the model using CE loss for the first 10 epochs. Then, we apply peer loss for the rest 10 epochs and the learning rate we use for these 10 epochs is 1e-6. The hyper-parameters for \textbf{$f$-div} is the same as those of peer loss. For \textbf{divide-mix}, we use the default hyper-parameters in the original paper. For \textbf{Jocor}, the hyper-parameters we use is as follows. The learning rate is 0.0001. $\lambda$ is 0.3. The epoch when the decay starts is 5. The hyper-parameters of \textbf{co-teaching} is similar to Jocor. For logitclip, $\tau$ is 1.5. For \textbf{taylorCE}, the hyper-parameter is the same as the original paper.

\subsection{Class-imbalanced learning benchmark}
\label{sec:class_learning}
The platform we use is the same as label noise detection. The backbone model we use is ResNet-50 \cite{he2016deep}. For different imbalance ratio ($\rho=10, 50, 100$). The class distribution is shown in Table \ref{tab:class_imbalance}. For all the methods, the base learning rate is 0.0001 and the batch size is 448. The dataset we use is not full dataset because we want to disentangle the noisy label and class imbalance learning. We use Docta and a pre-trained model trained with cross-entropy to filter the data whose prediction confidence is low. Due to the memorization effect, we fine-tune the model for 2 epochs to filter the data. We remove 45.15\% data in total where Docta removes 26.36\% while CE removes 25.00\% with a overlap of 6.20\%. Thus, the datset we use for class-imbalance learning is 54.85\% of the full dataset.

\section{Demo application of ADC in other fields} \label{app:demo_applicaiton}
Our Automated Dataset Construction (ADC) pipeline is best suited for image classification tasks where the relevant knowledge can be easily searched and retrieved from the internet. Example applications include, but are not limited to:
\begin{itemize}
 \item Food classification 
 \item Hairstyle classification
 \item Vehicle classification
 \item Home decor classification
 \item Plant classification
 \item Sport equipment classification
 \item Jewelry classification
\end{itemize}

\textbf{Food Classification} To illustrate the effectiveness of our ADC pipeline, let's consider a more detailed example of food classification. We used the prompt "Food Classification: Create a dataset with various types of cuisine, and sub-classes for specific dishes, ingredients, or cooking methods. Help me to find 10 different attributes to describe food." LLM generated a range of subcategories to describe different types of food, including, but are not limited to:

\begin{itemize}
    \item Cuisine type (Italian, Chinese, Indian, etc.)
    \item Dish Type (Appetizer, main course, dessert, etc.)
    \item Protein source (Beef, Chicken, Tofu, etc.)
    \item Cooking method (Grilled, Baked, Fried, etc.)
    \item Spice level (Mild, Medium, Spicy, etc)
    \item Allergen warning (Gluten-free, Nut-free, Dairy-free, etc.)
    \item Texture (Crunchy, Chewy, Smooth, etc)
\end{itemize}

Please feel free to use the prompt on your favorite LLMs, or modify it slightly for other tasks that interest you more. We tried various LLM versions from OpenAI, Meta, Google, and Claude, and all of them are competent to solve this task, albeit with different preferences for suggesting labels and descriptions.

\section{Copyright issue}

One possible approach to mitigate the potential copyright issues is to rely on the advanced features in search engines provided by the leading industry companies. For example, we can use the "Advanced Image Search => usage rights" function in Google Image Search, which allows users to filter search results by usage rights. 

However, We must clarify that our pipeline is provided "as-is" and that users are responsible for using the collected data at their own risk. We cannot guarantee that the data is free from copyright issues, and users must take their own steps to ensure compliance with applicable laws and regulations. This approach is similar to that taken by the LAION-5B dataset \cite{schuhmann2022laion}, which states that "The images are under their copyright."


\newpage

\end{document}